\documentclass[conference]{IEEEtran}
\IEEEoverridecommandlockouts
\usepackage{cite}
\usepackage{amsmath,amssymb,amsfonts}
\usepackage{algorithmic}
\usepackage{graphicx}
\usepackage{textcomp}
\usepackage{xcolor}
\usepackage[ruled,linesnumbered]{algorithm2e}
\usepackage{hyperref} 
\usepackage{subfig}
\usepackage{multirow}
\usepackage{url}
\usepackage{booktabs}

\newcommand\projname{ROAM\xspace}

\def\BibTeX{{\rm B\kern-.05em{\sc i\kern-.025em b}\kern-.08em
    T\kern-.1667em\lower.7ex\hbox{E}\kern-.125emX}}
\begin{document}

\title{ROAM: memory-efficient large DNN training via optimized operator ordering and memory layout\\
}

\author{\IEEEauthorblockN{1\textsuperscript{st} Huiyao Shu}
\IEEEauthorblockA{\textit{South China University of Technology} \\
\textit{Alibaba Group}\\
cscsyyp@mail.scut.edu.cn}
\and
\IEEEauthorblockN{2\textsuperscript{nd} Ang Wang}
\IEEEauthorblockA{\textit{Alibaba Group} \\
wangang.wa@alibaba-inc.com}
\and
\IEEEauthorblockN{3\textsuperscript{rd} Ziji Shi}
\IEEEauthorblockA{\textit{National University of Singapore} \\
\textit{Alibaba Group}\\
zijishi@comp.nus.edu.sg}
\and
\IEEEauthorblockN{4\textsuperscript{th} Hanyu Zhao}
\IEEEauthorblockA{\textit{Alibaba Group} \\
zhaohanyu.zhy@alibaba-inc.com}
\and
\IEEEauthorblockN{5\textsuperscript{th} Yong Li}
\IEEEauthorblockA{\textit{Alibaba Group} \\
jiufeng.ly@alibaba-inc.com}
\and
\IEEEauthorblockN{6\textsuperscript{th} Lu Lu}
\IEEEauthorblockA{\textit{South China University of Technology} \\
lul@scut.edu.cn}
\and
}

\maketitle

\begin{abstract}

As deep learning models continue to increase in size, the memory requirements for training have surged. While high-level techniques like offloading, recomputation, and compression can alleviate memory pressure, they also introduce overheads. However, a memory-efficient execution plan that includes a reasonable operator execution order and tensor memory layout can significantly increase the models' memory efficiency and reduce overheads from high-level techniques.

In this paper, we propose \projname which operates on computation graph level to derive memory-efficient execution plan with optimized operator order and tensor memory layout for models.  We first propose sophisticated theories that carefully consider model structure and training memory load to support optimization for large complex graphs that have not been well supported in the past. An efficient tree-based algorithm is further proposed to search task divisions automatically, along with delivering high performance and effectiveness to solve the problem. Experiments show that \projname achieves a substantial memory reduction of 35.7\%, 13.3\%, and 27.2\% compared to Pytorch and two state-of-the-art methods and offers a remarkable $53.7\times$ speedup. The evaluation conducted on the expansive GPT2-XL further validates \projname's scalability.

\end{abstract}

\begin{IEEEkeywords}
Deep Learning System, Memory Optimization, Large Model Support, Memory Efficiency
\end{IEEEkeywords}

\section{Introduction}
\begin{figure}
    \centering
    \includegraphics[width=0.5\textwidth]{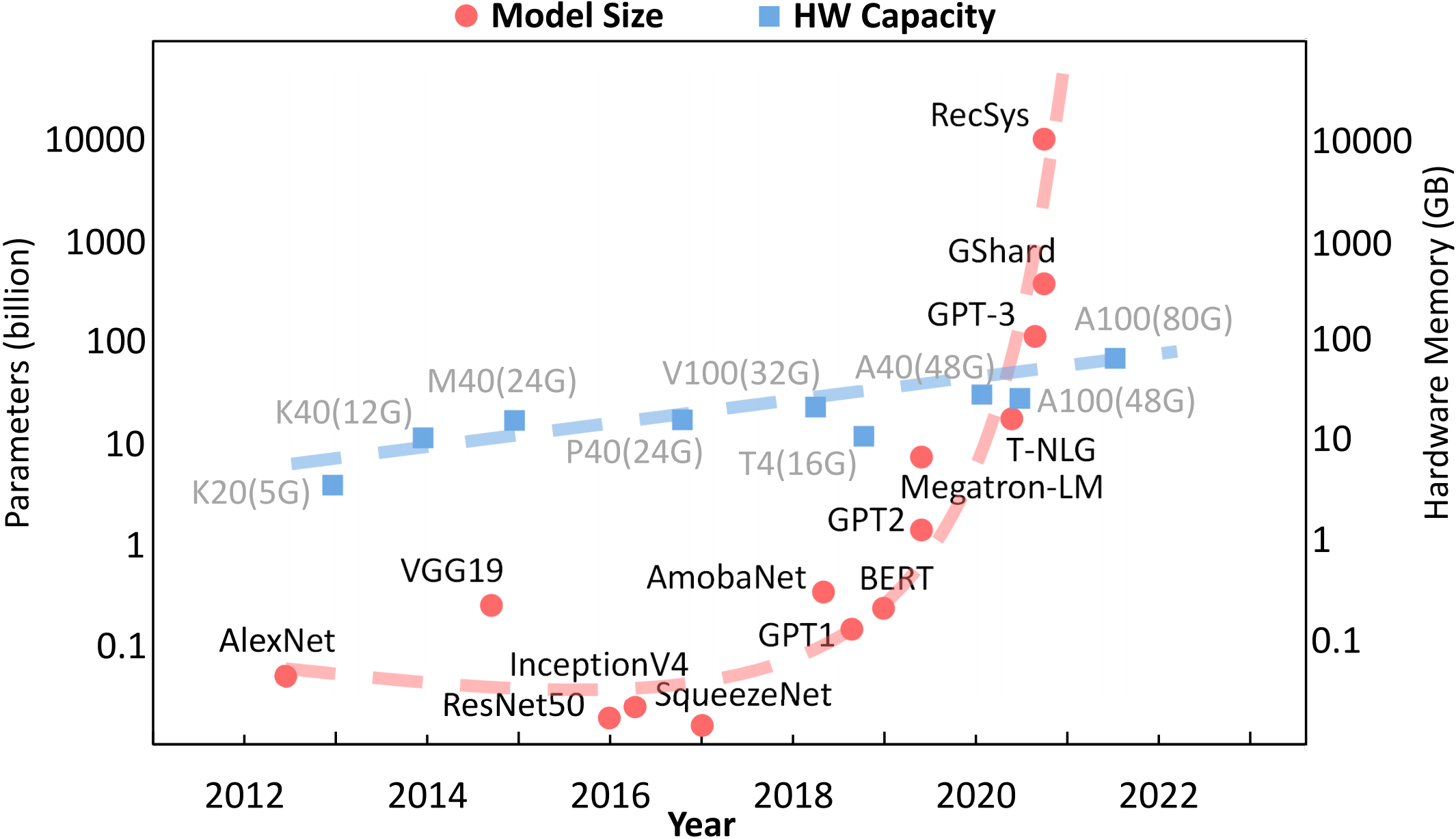}
    \caption{Recent advancements in model sizes have far surpassed the increment of accelerator memory capacity.}
    \label{fig:param_mem_trends}
\vspace{-0.5cm}
\end{figure}

Recently, many works \cite{Larger1,Larger2,Larger3} have suggested that employing a larger neural network constitutes an efficient way of improving accuracy. Consequently, the deep learning community has increasingly turned to larger DNNs to solve more complex tasks based on massive volumes of data, such as large-scale machine translation and high resolution image. 
Despite this trend, however, as shown in \autoref{fig:param_mem_trends}, the memory capacity of hardware has not kept pace with the growth of neural network size. Moreover, parameters only occupy a small fraction of the memory; during training, gradients, stashed activation, optimizer stats and framework workspace can significantly inflate the memory footprint. This highlights that the memory capacity is becoming the bottleneck in DNN training progress \cite{Harmony,MemBN2,MemBN3,LowMem}, limiting the exploration of more advanced DNN architectures. 

While distributed techniques enable the training of massive and complex models \cite{dist1,dist2,jia2022whale,shi2023tap}, they require massive expensive AI accelerators (e.g. GPUs) resources, limiting accessibility to a minority of individuals. Therefore, memory optimization for limited GPU resources remains an important issue. Many techniques have been proposed to relieve memory pressure. These include eviction and regeneration achieved by offloading \cite{ZeRO-Offload,flashneuron,caupchin,vDNN} or recomputation \cite{MemOptRec,AutoGC,SupNeu}, and quantization \cite{low-precision1,low-precision2,low-precision3} or compression \cite{COMET,Gist} techniques that involve representing tensors with fewer bits during training. Although effective in alleviating memory pressure, these approaches come with distinct disadvantages, such as time overhead and accuracy reduction.


\begin{figure*}[t]
    \centering
    \subfloat[]{
    \label{fig:simple_graph}\includegraphics[width=0.3\textwidth]{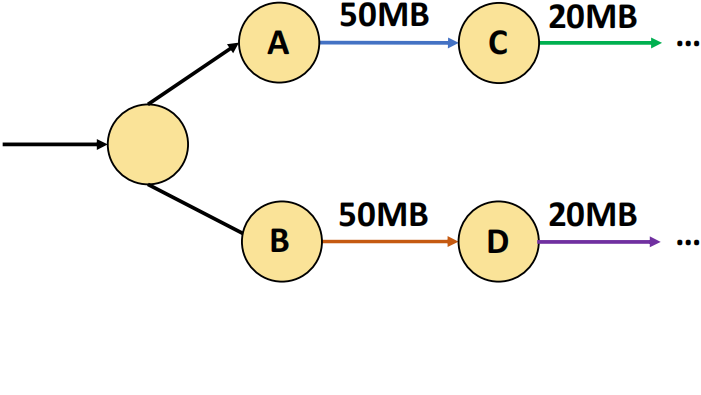}
    } 
    \subfloat[]{
        \label{fig:orders}\includegraphics[width=0.65\textwidth]{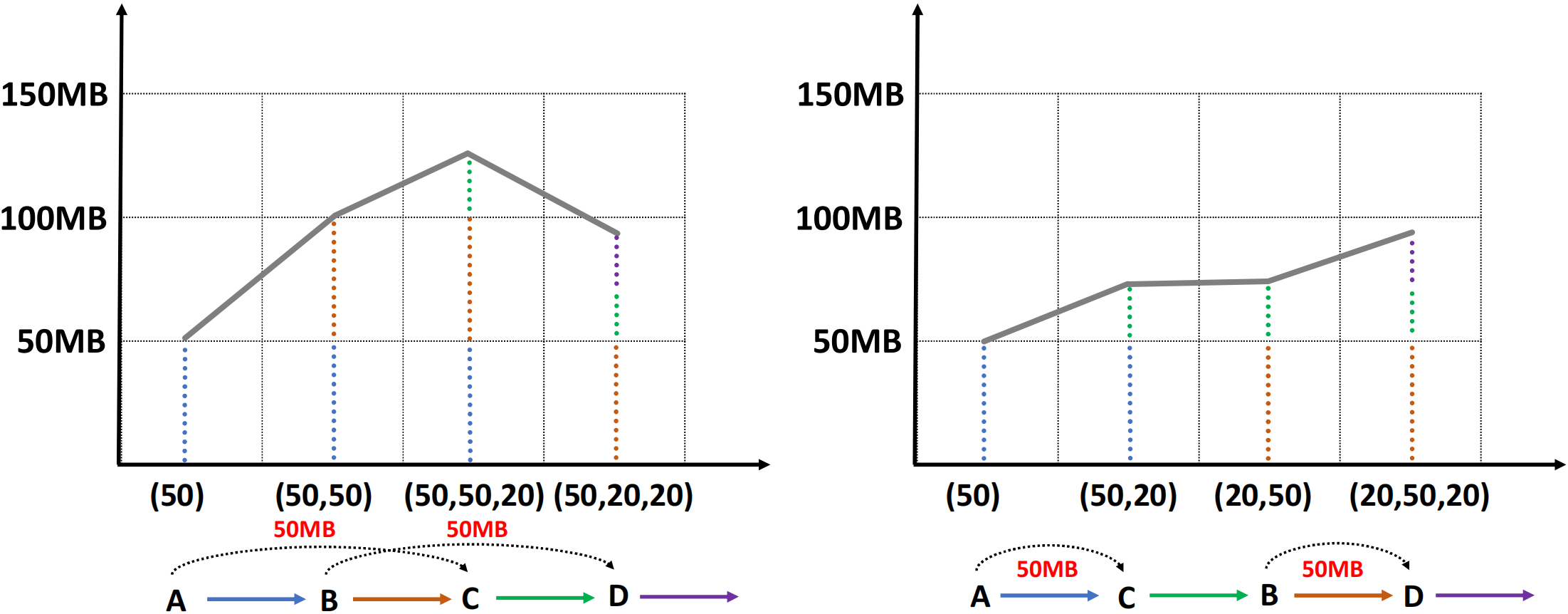}
    }
    \caption{Operator execution order theoretically affects the peak memory. The first execution order (A, B, C, D) results in both large tensors being preserved in memory simultaneously, leading to a peak memory of 120MB. The second order (A, C, B, D) prioritizes the execution of C, resulting in the early release of the large tensor and effectively reducing the peak memory to 90MB.}
     \label{fig:exe_order}
\vspace{-0.5cm}
\end{figure*}



These approaches are typically applied when models are not executed in a memory-efficient way, which might introduce unnecessary overhead that could be otherwise avoided with a more efficient execution plan. Our approach aims to improve the model's memory efficiency by reducing its memory requirements without performing additional tensor operations such as eviction and compression. We manage to reduce the theoretical peak memory of models and improve the memory reuse \cite{EfficientMMInf} efficiency. Current Deep Learning (DL) compilers \cite{TVM} and frameworks \cite{Pytorch,Tensorflow} rely on basic topological ordering algorithms that are oblivious to peak memory usage. Pytorch \cite{Pytorch} executes operators in the order they are defined in the program. Tensorflow \cite{Tensorflow} keeps a queue of ready operators and executes them according to the in-queue time. However, these execution orders are generally not memory-efficient. As shown in \autoref{fig:exe_order}, an inefficient execution order may have an operator schedule that emits large tensors simultaneously, resulting in high peak memory usage. 
\begin{figure}
    \centering
    \includegraphics[width=0.5\textwidth]{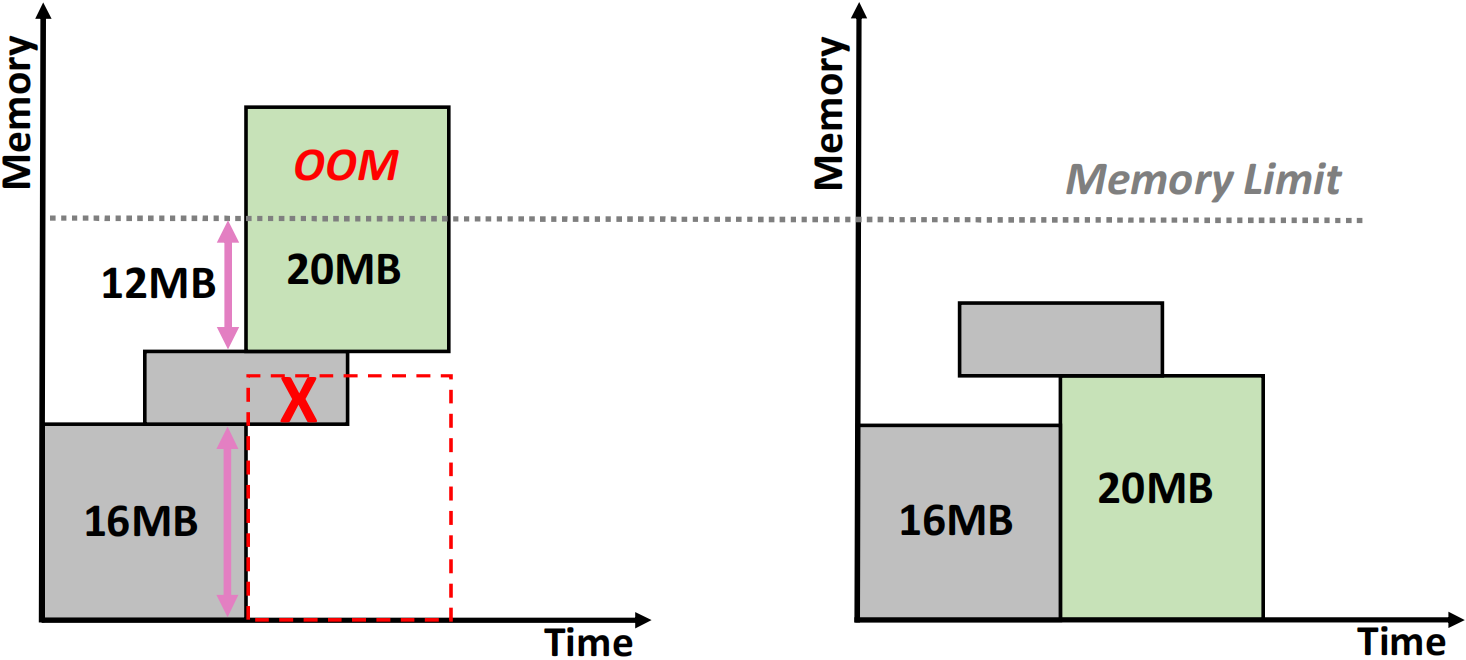}
    \caption{The memory layout of tensors also influences the actual peak memory. The left layout determines the tensors' address when they are created, resulting in OOM when the tensor with size 20MB is created, despite sufficient available space. In contrast, the right layout fully considers tensors lifetime and sizes, and enables memory reuse between tensors (16MB and 20MB), reducing the peak memory.}
    \label{fig:mem_layout}
\vspace{-0.5cm}
\end{figure}

Furthermore, inefficiencies in memory layouts can impose a significant influence on actual memory requirements. Improper memory layout of tensors results in low memory reuse efficiency, leading to data fragmentation between two adjacent tensors in memory, as shown in \autoref{fig:mem_layout}. Existing DL frameworks usually search for sufficiently large blocks of free memory in memory pools at runtime, or opt to allocate the necessary memory from physical sources. They decide the memory offsets of tensors only considering the generation time. However, memory reuse is also related to tensors' sizes and lifetime \cite{Tela}. Therefore, such a runtime allocation method is difficult to fully reduce the memory requirements by memory reuse. As a result, fragmented memory frequently occurs during the process of dynamic allocation due to variations in the lifetime and size between different tensors, which may lead to memory allocation failure due to lack of contiguous memory \cite{ZeRoMem}. 

In this work, we propose \projname, which operates on the computation graph level and seeks to optimize the execution plan for models and improve memory efficiency. We first propose sophisticated theories that jointly consider the characteristics of model structure and training memory load, and derive a memory-efficient execution plan for a large model from its divisions. Moreover, we overcome the ineffectiveness that divisions may introduce and then propose an efficient tree-based algorithm that automatically converts the overall large task into several smaller ones, enabling parallel implementation and optimization using the high-complexity but accurate methods (e.g., Integer Linear Program, ILP),  delivering high performance and effectiveness. Compared to the existing graph-level works, \projname can effectively and efficiently optimize large and complex training graphs that have not been well supported. Compared to high-level mitigation approaches, \projname requires no modification to the training procedure and introduces no additional overhead. Meanwhile, \projname is orthogonal to the high-level optimization techniques (e.g., offloading, computation), and can be used together. In summary, this paper makes the following contributions: 


\begin{itemize}
    \item We propose an effective framework \projname to optimize the execution plans for models that significantly improve the model's memory efficiency with no modifications to training procedures and without introducing additional overhead.
    \item We demonstrate the sophisticated theories that make it possible to derive effective memory-efficient execution plans for large complex computation graphs of training.  
    \item We provide an efficient algorithm that automatically converts the
overall large task into several smaller ones, enabling parallel implementation and optimization using high-complexity but accurate methods, delivering high performance and effectiveness. 
    \item We evaluate on a wide range of DNN models. Experiments show that \projname achieves a substantial memory reduction of 35.7\%, 13.3\%, and 27.2\% compared to Pytorch and two state-of-the-art methods and offers a remarkable $53.7\times$ speedup. The evaluation conducted on the expansive GPT2-XL further validates \projname’s scalability.

\end{itemize}

\section{Motivation}

With the prevalence of large models, distributed training has become a vital technique that requires massive and expensive AI accelerators (e.g. GPUs). Low memory efficiency during distributed training can increase demands on GPUs, hiking up training expenses. On the other hand, tensor-wise mitigation measures (e.g. Offloading, recomputation) running when models may introduce unnecessary overheads that could be decreased by improving models' memory efficiency. 


The memory efficiency of models is determined by two aspects: 
\begin{itemize}
    \item Theoretical peak memory, which is determined by the execution order of operators;
    \item Efficiency of memory layout, which is reflected as fragmentation at runtime. Ideally, an optimal memory layout can achieve an actual peak memory that closely approximates the theoretical peak memory.
\end{itemize}


While improving memory efficiency can reduce the memory requirements of models remarkably as exhibited  in \autoref{fig:exe_order}
 and \autoref{fig:mem_layout}, current widely used approaches are inadequate for dealing with large models in training due to poor scalability or insufficient optimization. 

\textbf{The Recognized NP-Hardness of the Problem.} Previous works \cite{OrderNPC1, OrderNPC2, DSA-NP,2SP} prove that the optimal scheduling for DAGs and memory layout optimization, which is also known as the Dynamic Storage Allocation (DSA) problem, are typical NP-Complete and NP-hard problems, respectively. Therefore, it is challenging to find the optimal solution in polynomial time.

\textbf{Large Scale Graph and High Flexibility in Training.} Above two problems are often transformed to an integer linear programming program, which can be solved near-optimally\cite{ILP1,ILP2,ILP3}. However, in practice, the efficiency of this method can be hindered by the scale of computation graphs. For example, optimizing VIT \cite{VIT} with Adam optimizer \cite{Adam} can take more than 60 minutes. We counted the number of operators in the test computation graph, which was only around 2000. However, the training computation graph of recent prevailing large models such as GPT2-XL\cite{GPT2-XL} consists of more than 10,000 operators, making it difficult to solve directly using ILP. Another problem is the high flexibility of the weight update operations, which can be scheduled immediately after gradients generation or at any later timestep. Such flexibility also increases the complexity of ILP problems a lot.



\textbf{Rich memory reuse pattern.} The high complexity of the ILP approach on large graphs introduces substantial time cost, making developers resort to greedy strategies for more efficient search. However, the memory reuse pattern between tensors, especially those with similar lifetimes and sizes, can be very rich and complicated, which makes the performance of greedy strategies sub-optimal. For example,  an existing state-of-art approach \cite{LLFB} that takes tensors' lifetime and best-fit strategy into consideration to make assignments for tensors has been proven to be effective. This approach selects the lowest available offset and places the tensor with the longest lifetime on it. However, although they can deal with the memory reuse between tensors with significantly different lifetimes, they fail to optimize tensors with similar lifetimes. 




\section{Prliminary}

\subsection{Memory Loads During Training}
There are mainly three stages in training: forward propagation pass, backward propagation pass, and weight update pass. During training, the majority of tensors generated in the forward pass must be retained in memory until their corresponding gradients are calculated.
Based on the lifespan of these intermediate tensors, we categorize the\textbf{\textit{ tensors that are created in forward pass and preserved in memory to calculate gradients in backward pass as activations and the others as temporary buffers}}. During the backward pass, once the gradients are ready, new weight parameters can be calculated as defined by the optimizers. The memory footprint in this stage highly depends on the optimizer: if the optimizer relies on no other information (e.g., SGD \cite{sgd}), it will not consume much memory; however, if the optimizer is relatively complex by tracking other information like exponential moving averages (e.g., Adam), there could be numerous temporary buffers created in this process, making it heavier.

Typically, the memory usage of a neural network escalates during the forward pass and gradually decreases during the backward pass \cite{AutoGC, TrainMem12, TrainMem13}. As a result, the theoretical peak memory usage is likely to occur around the loss calculation operations. However, the situation is not always this straightforward, as some temporary buffers, which can be substantially larger than activations, have a significant influence on overall memory usage.

\subsection{DNN Graph}
We represent the computation graph of a DNN as a directed acyclic graph G = (V, E) with $V=\{v_{i}\}, (i=1,2,...,n)$ and $E=\{e_{i}\}, (i=1,2,...,k)$ as the vertices and edges in the graph, respectively. $n$ and $k$ denote the number of vertices and edges, respectively. The vertices represent DNN operations, such as convolution, and matrix multiplication, while edges represent tensors exchanged by operators. The size of tensor $e$ is denoted as $size_{e}$. 

We model the execution time of a single operator as a discrete timestep. The set of all possible operator execution orders of an arbitrary graph is represented by $S$.  We define the function $Tp(\cdot)$ as the measure of \textbf{T}heoretical \textbf{p}eak memory. $Tp(v_i)$ is calculated as the sum of sizes of tensors that are alive when operator $v_i$ is scheduled. $Tp(G, s)$ can be calculated by recording the peak memory that operators need to request to run the graph with a specific order $s$.  We define the set of all possible memory layouts of tensors in an arbitrary graph as $M$. For each $m \in M$, it contains the memory offsets of all corresponding tensors, which can be represented as $m[e]$ for a specific tensor. $Peak(m(G))$ denotes the actual peak memory for the graph with tensors assigned as offsets in a specific memory layout $m$. 

\section{Methodology}
\begin{figure}[t]
    \centering
    \includegraphics[width=0.5\textwidth]{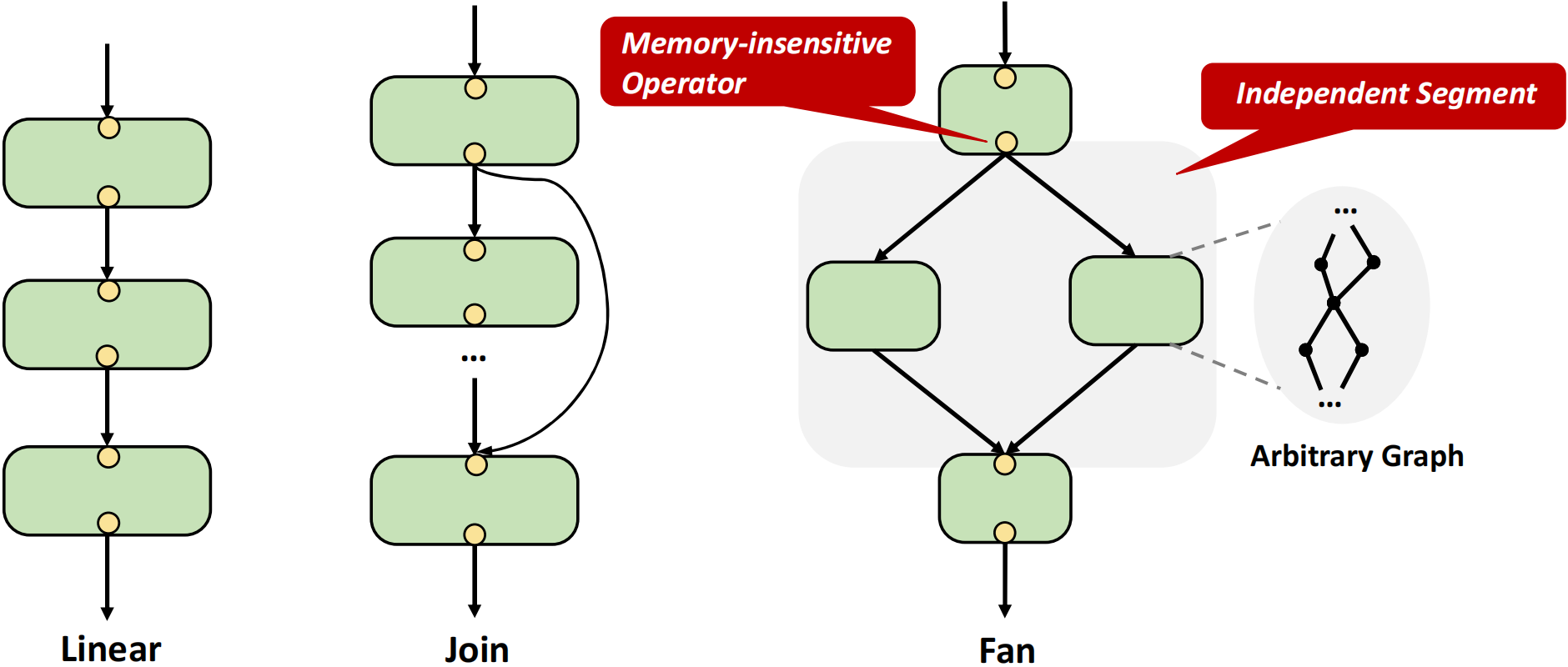}
    \caption{The common structures in models. There are common memory-insensitive operators that can break the graph into independent segments. Each independent segment contains an arbitrary graph.}
    \label{fig:structure}
\vspace{-0.5cm}
\end{figure}

In this section, we present the theories for optimizing both theoretical peak memory and actual memory requirements. To optimize theoretical peak memory, we first introduce basic definitions and formulate the optimization problem for solving a memory-aware execution order. We start with simple computation graphs and progress to complex training graphs. To reduce actual memory requirements, we analyze and overcome the difficulty of applying divide-and-conquer theories in memory layout optimization and formulate the optimization objective. We then present a tree-based algorithm that describes how to search for divisions, enabling efficient and effective solving of the overall optimization task with accurate methods. Finally, we introduce ILP, an accurate approach applied in our framework.

\subsection{Operator Ordering Optimization}

Computation graphs are decided by the model definitions in program and running mode (inference or training) \cite{graph_ir}. While coarse layers are transformed into one or multiple fine-grained operators, computation graphs preserve the characteristics of the model structure. \autoref{fig:structure} shows the common structures within current prevailing models \cite{SupNeu}. 
 
We have observed that some operators have a fixed scheduling timestep throughout the entire computation graph. Such operators are defined as \textbf{Memory-Insensitive Operators} in this paper. Operators situated between two memory-insensitive operators can be scheduled with numerical topological orders that result in different peak memory. For computation graphs that have simple linear structures, with all inputs and outputs fed in and generated at the same time, that are common in inference, memory-insensitive operators naturally break the whole computation graph into different segments as shown in \autoref{fig:structure}. We refer to these segments as \textbf{Independent Segments (IS)}. Specifically, an independent segment is denoted as $(v_i, v_j)$ with $v_i$ and $v_j$ as the boundary memory-insensitive operators.

\textbf{Operators Order Optimization based on IS.} For a graph that is broken into independent segments (\{$IS_0$, $IS_1$, ..., $IS_n$\}), its theoretical peak memory is decided by the peak memory of independent segments and the memory requirement of the boundary operators. Therefore, the theoretical peak memory of the whole graph $Tp(G,s)$ can be formulated as follows:
\begin{equation}
 \forall s \in S \ \   Tp(G, s) = \mathop{\max}(
        \mathop{\max}_{i} Tp(IS_i, s_i), 
        \mathop{\max}_{j} Tp(v_j))
\end{equation}
where $v_j$ represents the memory-insensitive operator that serves as a boundary between independent segments. Since the execution order between memory-insensitive operators and their surroundings is clear,  $Tp(v_j)$ is constant, equaling the sum of sizes of tensors that are created earlier but are not freed. Therefore, we can optimize the theoretical peak memory of the entire graph by separately optimizing for independent segments. The optimization objective can be defined as follows:
\begin{equation}
  \forall i \in \{1, ... ,n\} \quad   \mathop{\arg\min}_{s_i \in S_i} Tp(IS_i, s_i)
\end{equation}
By solving the objective for all independent segments, all low-peak memory execution sub-orders are generated and the overall execution order for the whole graph can be derived as:
\begin{equation}
    s = [s_0, s_1, ..., s_n]
\end{equation}

\textbf{Challenges in Training Computation Graph.} As mentioned earlier, there are three stages in training: forward propagation pass, backward propagation pass, and weight update pass \cite{MODeL}. We can find memory-insensitive operators in the backward pass that correspond to memory-insensitive operators in the forward pass \cite{AutoGC}. The graph of forward and backward propagation passes can be broken into independent segments. 
However, a significant challenge arises concerning the weight update operation, which has not been memory-efficiently addressed in previous works.

\begin{figure}
  \centering
  \includegraphics[width=0.9\linewidth]{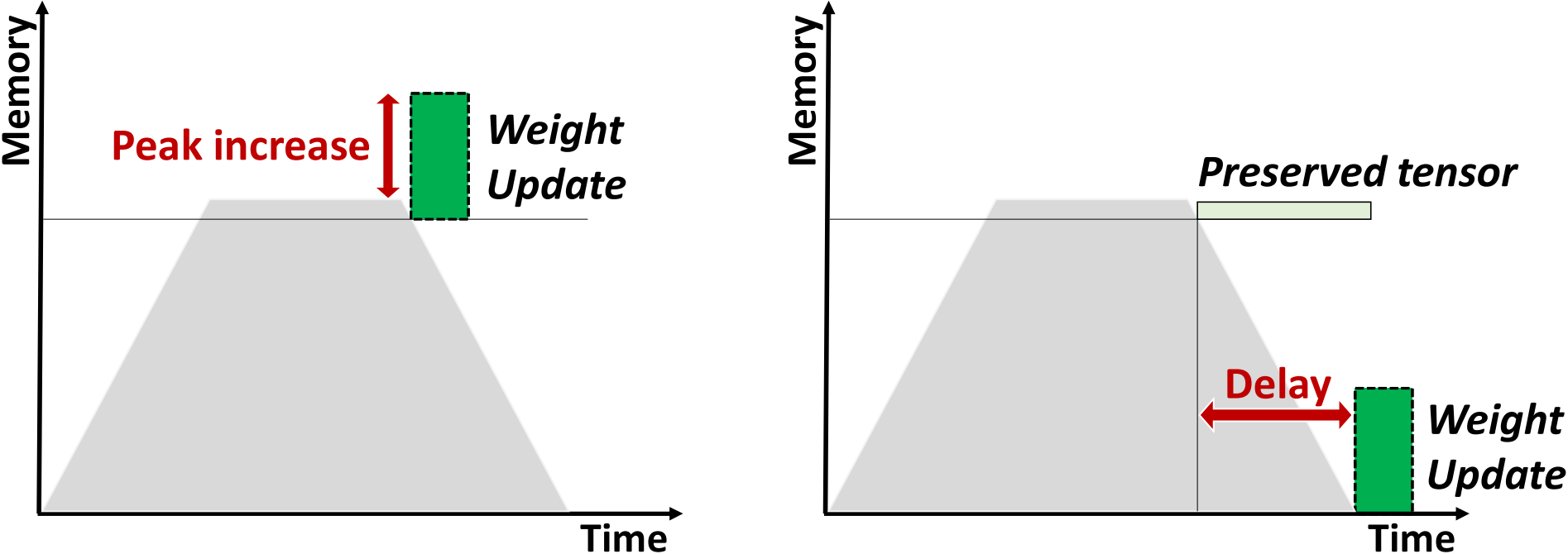}
  \caption{(a) demonstrates executing the weight update operation immediately may increase memory pressure. (b) demonstrates delaying such weight update operation reduces peak memory (b).}
  \label{fig:weight_update}
\end{figure}

Once the gradients are generated, the corresponding weight update operations are ready to be scheduled. Thus, the scheduling of weight update operations demonstrates strong flexibility. The scheduling timestep of the weight update operation can greatly influence the peak memory. Two extreme cases are considered, scheduling weight update immediately when the gradient is generated or scheduling all weight update operations after finishing the backward propagation pass. As shown in \autoref{fig:weight_update}, executing weight updates as early as best may produce many large temporary buffers when memory consumption is intensive, for example, when most activation tensors are preserved in memory, leading to higher memory pressure.  Nevertheless, choosing to delay all the weight update operations is not a preferable proposition since every gradient must be preserved for a substantially long period of time. Therefore, it is important to assign the weight update operations into proper independent segments to be scheduled. 

\begin{figure}[t]
    \centering
    \subfloat[]{
        \label{fig:LLF_sample}
        \includegraphics[width=0.5\textwidth]{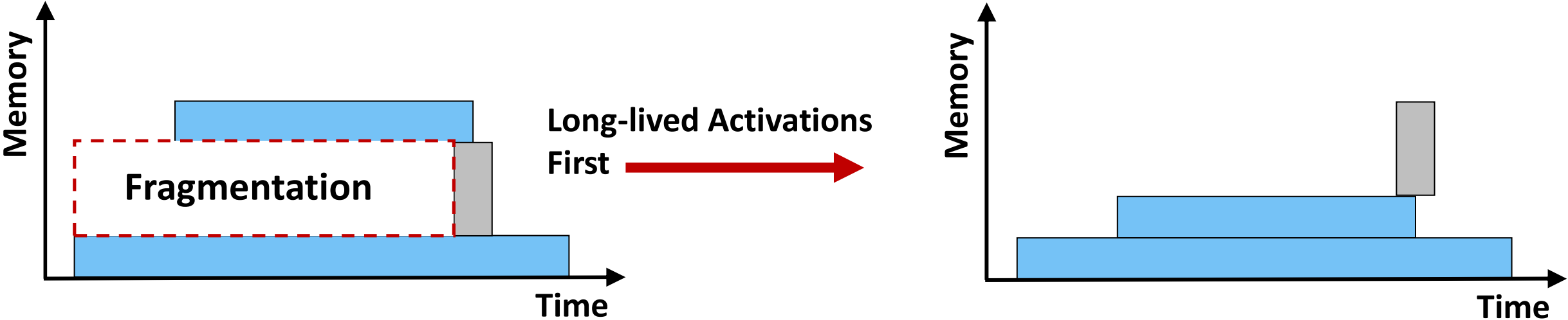}
    } \newline
    \subfloat[]{
        \label{fig:concate_sample}\includegraphics[width=0.5\textwidth]{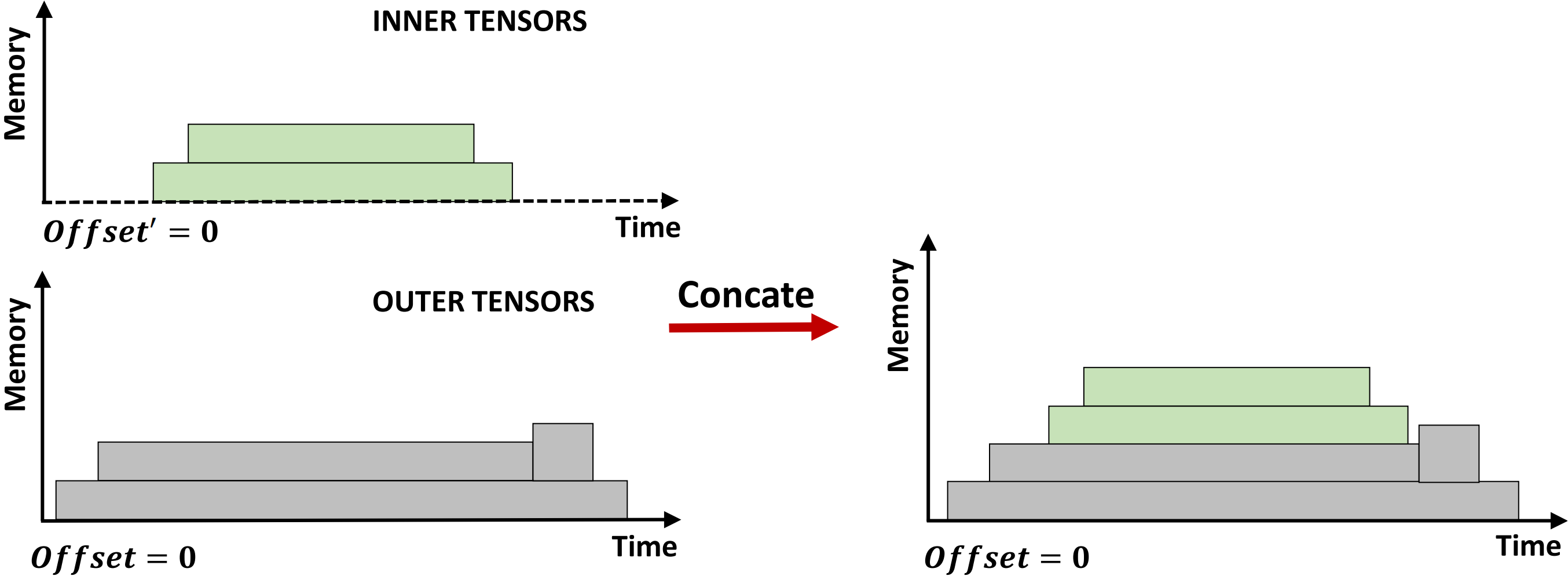}
    }
    \caption{Long-lived activations first  force the activations placed at the bottom to eliminate long-term fragmentation (a), thus making it possible to merge the sub-memory-spaces like (b).}
    \label{fig:LLF_concate}
\end{figure}

\textbf{Memory-aware Scheduler for Weight updates.} To mitigate the abovementioned issue, we propose the following strategy to decide whether to schedule the weight update operations in the current segment or later. The first step of this method involves calculating the sum of the sizes of activations to estimate the peak memory usage:
\begin{equation}
    esti\_pm = \mathop{\sum}\limits_{e \in E_{atvs}} size_e
\end{equation}
where $E_{atvs}$ represents the set of activations. Notice that $esti\_pm$ is just an estimated value. Then the memory consumed by activations when scheduling the weight update operation immediately at timestep $t$ is approximated as: 
\begin{equation}
    mem\_atvs_{t} = \mathop{\sum}_{e \in E_{atvs}} is\_alive_{e,t} \cdot size_{e}
\end{equation}
where $is\_alive_{e,t}$ equals 1 when tensor may be alive at timestep t and equals 0 if not. The $is\_alive$ values are derived from the earliest possible execution time and the latest mandatory execution time of operators, which calculates the number of all transitive predecessors and successors, respectively.


Then the memory usage of executing the weight update operation at $t$ can be estimated as:
\begin{equation}
    mem\_used_{t} = mem\_atvs_{t} + \alpha \cdot size_{grad}
\end{equation}
where $size_{grad}$ denotes the size of gradients in the weight update branch. $\alpha$ is a coefficient pre-determined by the optimizer type, which can be derived from the memory requirement of tensors in the weight update operation as shown in \autoref{fig:weight_update}. When using the Adam optimizer, the tensors in the weight update branches can be placed with three layers, thus $\alpha$ equals 3 for Adam.

A new problem arises as a result of the delayed weight update: it prolongs the lifespan of tensors. To mitigate it, we introduce \textit{delay radius} $r$ to constrain the delaying operation. $r$ is determined empirically according to the tensor sizes in neural networks. We observe that the benefit of delaying operation outweighs the cost of increased lifespan when the following conditions are met:

\begin{itemize}
    \item the ratio between the size of the tensor in the weight update branch and the average size of the tensor is greater than $r$; and
    \item the estimated $mem\_used_t$ is greater than $esti\_peak$;
\end{itemize}

Based on the above strategy, we can assign weight update operations into proper independent segments to be optimized. Thus the theoretical peak memory optimization for training computation graphs can be solved with equation (2) and the overall optimized execution order can be derived with (3).

\subsection{Memory Layout Management}

While the operator scheduler optimizes the theoretical peak memory, memory layout also has a significant impact on memory usage. Specifically, memory layout optimization refers to the arrangement of memory offsets for a set of tensors with their lifetime and sizes. Existing works\cite{EfficientMMInf,han2006buffer} mainly focus on small programs. However, tensors in training exhibit the characteristics of being massive in number and diverse in lifetime, which makes layout optimization especially challenging. 

\begin{figure}
    \centering
    \includegraphics[width=0.45\textwidth]{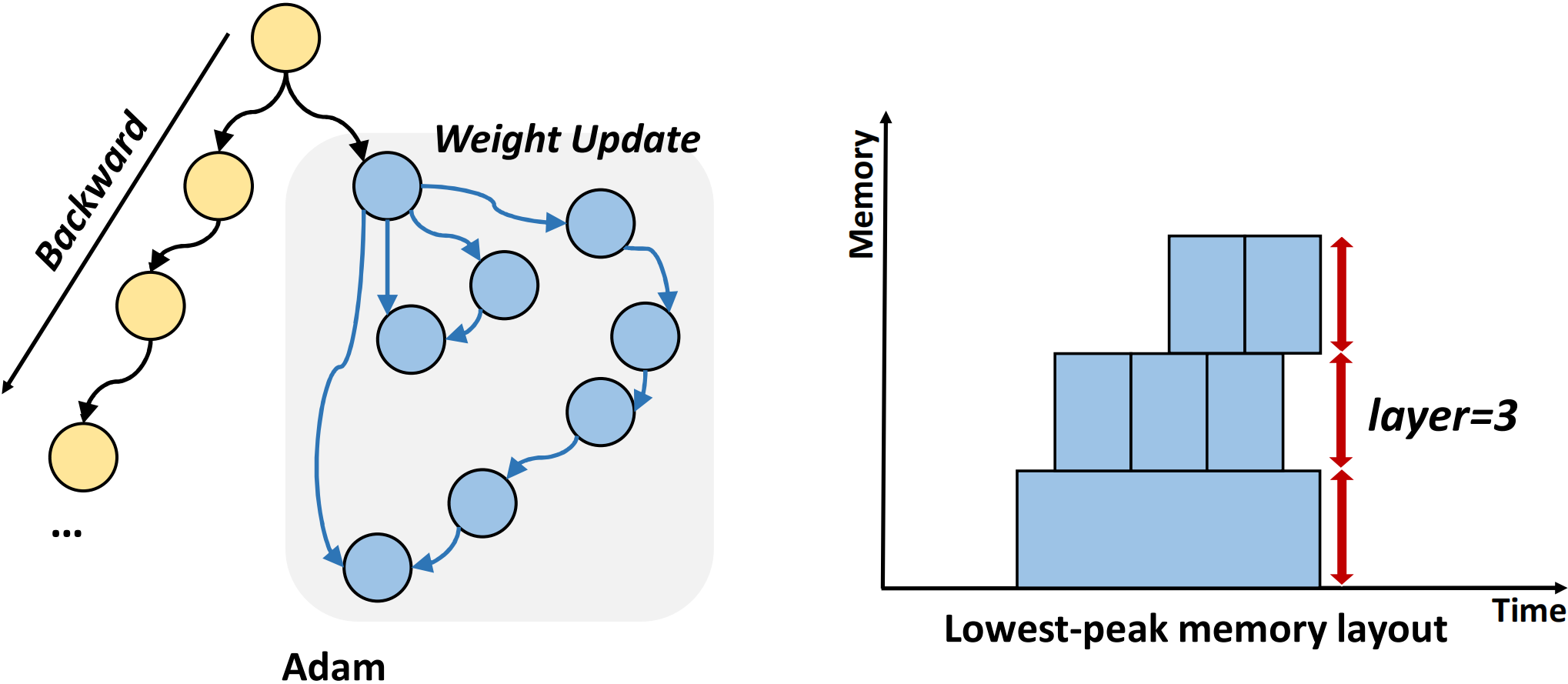}
    \caption{(a) shows the structure of the Adam optimizer in the computational graph, which produces multiple temporary buffers during weight update. (b) shows the lowest-peak memory layout of temporary buffers of tensors in the weight update branch, which takes 3 layers to store all tensors by memory reusing.}
    \label{fig:wu_alpha}
\end{figure}

\textbf{Objective Definition.} To tackle such challenges, we aim to explore ways to simplify the optimization complexity from a graph structure perspective. Similar to equation (1) in theoretical peak memory optimization, we attempt to derive a reasonable solution for actual peak memory as follows:
\begin{equation}
\begin{aligned}
     \forall m \in M \quad Peak[m(G)] = & Peak[m_0(G_{0}^{'}) + m_1(G_{1}^{'}) + \\ 
    & ... + m_n(G_{n}^{'})]
\end{aligned}
\end{equation}
where $G^{'}$ is the set of subgraphs that splits the whole graph. The '+' symbol represents the concatenation operation between different memory layouts. 
The challenge lies in:
\begin{itemize}
    \item How to decide the suitable subgraphs set of $\{G_{0}^{'}, G_{1}^{'}, ..., G_{n}^{'}\}$;
    \item How to effectively define the specific operation in '+'.
\end{itemize}
 Both of these points simultaneously affect the efficiency of equation (7).

\textbf{Constraints for Subgraph.} Tensors' lifetime and sizes jointly guide the optimization of memory layouts\cite{EfficientMMInf}. Ideally, it is necessary to optimize all tensors that overlap in lifetime together to maximize memory reuse between them. As mentioned earlier,  the independent segment ($v_{i}, v_{j}$) contains all the operators that can be scheduled between $t_{v_{i}}$ and $t_{v_{j}}$.  That means all the tensors that have intersecting lifetimes within $[t_{v_{i}}, t_{v_{j}})$ can be gathered together. Since most activations should be preserved in memory until they are consumed in the backward pass, $G_{i}^{'}$ should keep the producers in the forward pass and consumers in the backward pass to express the full lifetime of activation tensors. Therefore, we formulate $G^{'}$ with independent segments in the forward pass and the corresponding independent segments in the backward, gathering tensors with overlapping lifetimes as much as possible. 

However, since the training graphs always have complex structures, it is difficult to guarantee that all tensors are created and freed in the same subgraph. To mitigate the potential ineffectiveness introduced by such tensors, we propose an efficient strategy to deal with the shared tensors whose lifetime extends across different subgraphs $G^{'}$ as follows. 

Considering that repeating optimization for shared tensors leads to one-tensor-multiple-addresses and makes it difficult to concatenate memory layout, we make decisions for each tensor regarding which subgraph it should be optimized in according to the memory consumption characteristics.

Shared tensors can be divided into four types in the subgraphs that overlap with their lifetime:
 \begin{itemize}
    \item \textit{CIFO},  \textbf{C}reated \textbf{I}n but \textbf{F}reed \textbf{O}utside the subgraph. 
    \item  \textit{COFI},  \textbf{C}reated \textbf{O}utside but \textbf{F}ree \textbf{I}nside the subgraph.
    \item  \textit{COFO}, \textbf{C}reated and \textbf{F}reed \textbf{O}utside the subgraph.
\end{itemize}
A shared tensor must be CIFO in a subgraph and COFI in another subgraph. 

\begin{figure}
  \centering
  \includegraphics[width=0.85\linewidth]{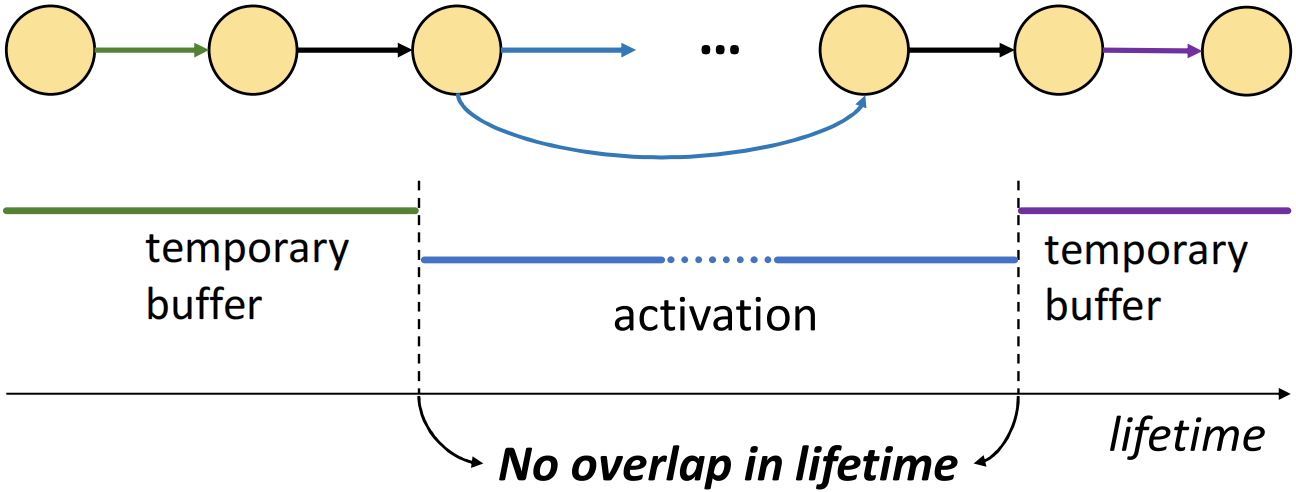}
  \caption{Activations can reclaim the memory space of temporary buffers when they are freed in forward. Similarly, temporary buffers created in backward can reuse activation's memory space.}
  \label{fig:atv_reuse}
\vspace{-0.5cm}
\end{figure}

As shown in \autoref{fig:atv_reuse}, activations have the ability to reclaim memory space from temporary buffers once these buffers are released during the forward pass. Similarly, temporary buffers can efficiently reuse the memory space of activations after they are released during the backward pass. Consequently, shared tensors that serve as temporary buffers and are released during the forward pass should be optimized in the subgraph where they exhibit the COFI type. Conversely, shared tensors functioning as temporary buffers and generated during the backward pass should be optimized in the subgraph where they manifest the CIFO type.

As for activations, optimization should occur in the subgraph where they are freed. This decision is grounded in the fact that the vast majority of activations created during the forward pass are retained in memory until the corresponding gradient computation takes place, leading to a limited number of temporary buffers during the forward pass. By optimizing the activations in the subgraph where they demonstrate the COFI type, a high level of memory reuse efficiency is achieved between activations and temporary buffers created during the backward pass.

On the other hand, shared tensors in a subgraph exhibiting the COFO type need not be optimized. This is due to the absence of any possibility for memory reuse between these shared tensors and the inner tensors. As a result, optimization efforts are best focused on the aforementioned cases of COFI and CIFO types for shared tensors.

\textbf{Constraints for Single Memory Layouts and concatenating operation.} The concatenating operation, denoted by "+", involves the merging of tensors assigned in different memory layouts (memory spaces) into a unified memory space. Merely stacking one memory layout above or below the other does not ensure the efficiency of the resulting layout, even if the individual memory layouts are optimized at their core. The underlying issue lies in the potential long-term fragmentation that arises from the presence of temporary buffers and activations in separate memory layouts, as exemplified in Figure\autoref{fig:LLF_sample}. To address this concern, we impose constraints that enforce a continuous placement of activations at lower offsets, thereby preventing interleaving between activations and temporary buffers. The concatenating process, depicted in Figure\autoref{fig:concate_sample}, entails positioning the memory layout with a shorter lifespan atop the other layout. This approach utilizes the cumulative size of long-lived activations as the fundamental offset, effectively mitigating the long-term fragmentation during the concatenating process.

\begin{figure}[t]
  \centering
  \includegraphics[width=0.9\linewidth]{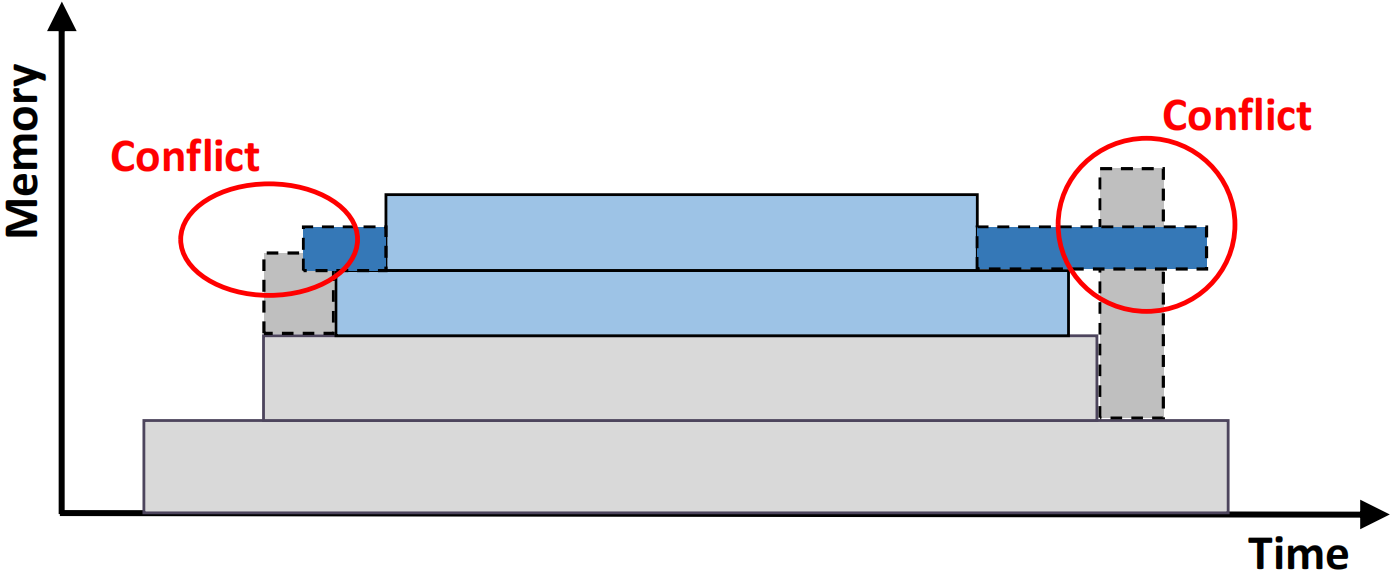}
  \caption{Address conflict between tensors.}
  \label{fig:locations_conflict}
  \vspace{-0.5cm}
\end{figure}

\textbf{Memory Layout Optimization Based on Subgraph.} Building upon the aforementioned constraint, the overarching optimization of memory requirements can be accomplished by independently optimizing sub-memory-layout. The optimization objective is meticulously defined as follows:

\begin{equation}
  \forall i \in \{1, ... ,n\} \quad  \mathop{\arg\min}_{m_i \in M_i} Peak[m_i(G_i^{'})]
\end{equation}
Supposing the lifetime of activations in $G^{'}$ with smaller subscript is shorter, the overall memory layout of the whole graph $m$ can be derived from the optimized results $\{m_0, m_1, ..., m_n\}$ as follows:

\begin{equation}\label{3}
    \left\{
        \begin{array}{lr}
                base_i = base_{i-1} + \mathop{\sum}\limits_{e \in m_{i-1}^{atvs}} size_e &  \\
                \forall e \in m_i, m_i[e] = base_i + m_i[e] & \\
        \end{array} 
    \right.
\end{equation}
 where $m_{i-1}^{atvs}$ indicates the set of activations in memory layout $m_{i-1}$.  Throughout the concatenating process, another concern arises regarding address conflicts, as illustrated in \autoref{fig:locations_conflict}, which may arise among tensors. To effectively address this issue, a strategic approach is employed. Temporary buffers characterized by smaller sizes and shorter lifetimes are selectively re-assigned after the completion of the concatenating operation. As the addresses of other tensors remain fixed, only a small subset of tensors requires reassignment, allowing for precise and optimized handling of their addresses through well-defined techniques.


\subsection{Subgraph Tree }
Guided by equations (2) and (7), we present a highly efficient framework designed to address the optimization challenge in large and intricate graphs. The entire graph is subdivided into more granular subgraphs, forming a subgraph tree that reflects these divisions. To achieve optimized memory layouts while factoring in tensor lifetimes comprehensively, we commence by seeking divisions in the entire graph that yield \textbf{I}ndependent sub\textbf{G}raphs (IG), wherein all tensors are created and freed exclusively within the subgraph. Recognizing that locating an independent subgraph of the appropriate scale may not always be feasible, we subsequently divide larger independent subgraphs into multiple \textbf{D}ependent sub\textbf{G}raphs (DG) to enable the application of accurate yet intricate methodologies efficiently.

\begin{figure}[t]
  \centering
  \includegraphics[width=\linewidth]{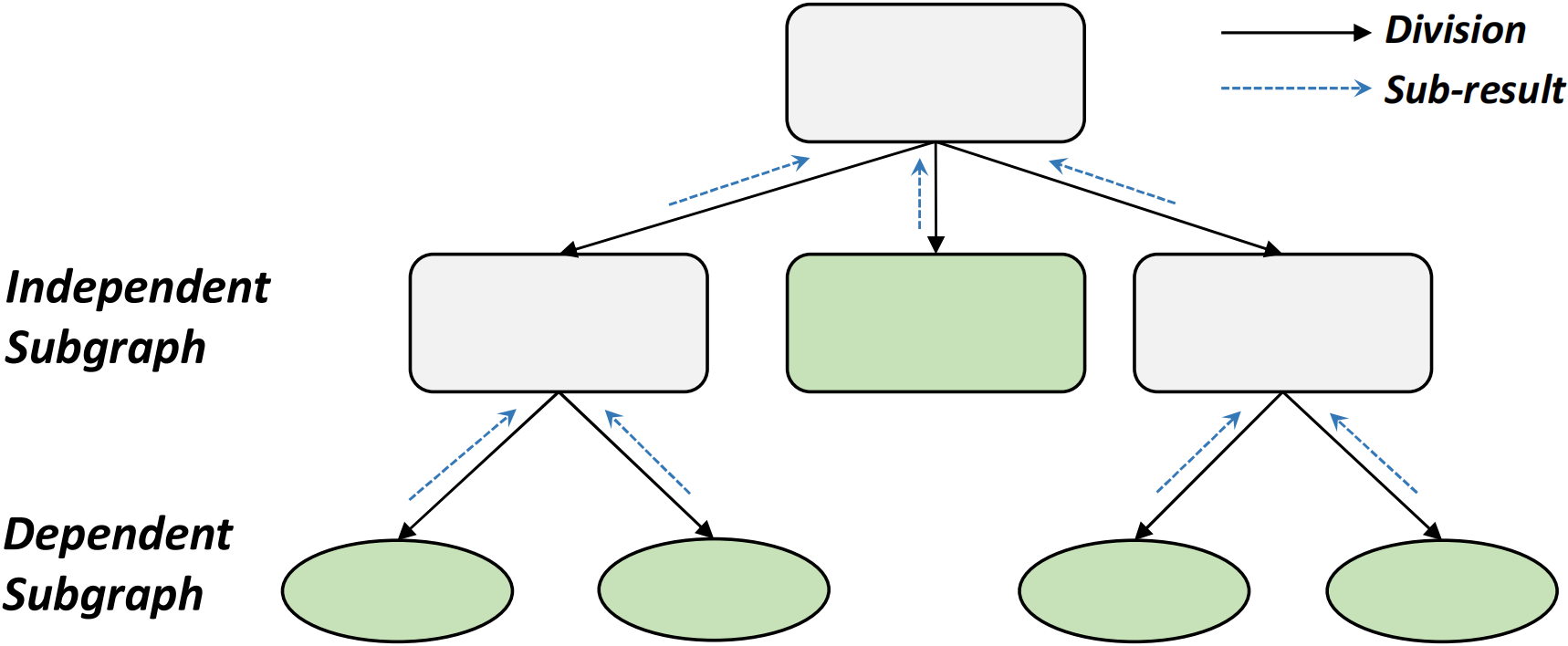}
  \caption{Subgraph Tree of a DNN graph. The original graph is first split into independent subgraphs. The large independent subgraphs are further split into dependent subgraphs.}
  \label{fig:task_tree}
  \vspace{-0.5cm}
\end{figure}

Each independent subgraph and dependent subgraph are then formatted with an independent segment from the forward pass, denoted as (outer\_fwd, inner\_fwd), and the corresponding independent segment from the backward pass, denoted as (inner\_bwd, outer\_bwd).




\textbf{Efficient Configuration.} In order to deliver high efficiency, all optimization tasks for subgraphs must adhere to an efficient configuration tailored to the specific approach employed. In the case of the ILP (Integer Linear Programming) technique utilized within our framework, we introduce a user-defined parameter, denoted as $node\_limit$, which serves to constrain the size of the graph, thereby preventing prolonged solving periods.

 
 The principal procedure of independent subgraph generation and split-down is delineated in \autoref{construct_ST}. During the process of independent subgraph generation, the inner boundaries (inner\_fwd and inner\_bwd) are initially set to $None$, while the outer boundaries (outer\_fwd and outer\_bwd) are initialized as the last and first memory-insensitive operators encountered in the forward and backward pass, respectively. Gradually, the search radius is expanded until two memory-insensitive operators capable of forming an independent subgraph with the inner boundaries are identified. Upon successful generation of an independent subgraph, the outer boundaries are then updated to serve as the inner boundaries for the subsequent iteration. In scenarios where the number of operators in the generated independent subgraphs surpasses the user-defined parameter \textit{node\_limit}, a splitting-down process for independent subgraphs is launched prior to proceeding to the next iteration. The process of generating dependent subgraphs mirrors that of independent subgraphs. Specifically, within a large independent subgraph, the memory-insensitive operators are traversed to generate dependent subgraphs with a reduced number of nodes, ensuring that the resulting subgraphs possess fewer nodes than the specified \textit{node\_limit}.
 

\begin{algorithm}
  \caption{ConstructSubgraphTree()}\label{construct_ST}
  \SetKwData{Left}{left}\SetKwData{This}{this}\SetKwData{Up}{up}
  \SetKwInOut{Input}{input}\SetKwInOut{Output}{output}

\Input{
    DNN graph $\Rightarrow$ G,  \\
    Memory insensitive operators: $\Rightarrow$ $V^{'}$: ($\mathbf{v_{1}^{'}}, \mathbf{v_{2}^{'}}\dots, \mathbf{v_{i}^{'}}$) \\
    The maximum number of  nodes in subgraphs: \\
    $\Rightarrow$ node\_limit
}

\Output{Subgraph tree $root$}

\BlankLine
$root \gets CreateSTNode(G)$ \\
$inner\_fwd, inner\_bwd \gets None, None$ \\
$outer\_fwd, outer\_bwd \gets lt\_fwd\_mi, ft\_bwd\_mi$ \\
\While{outer\_fwd and outer\_bwd}{
    $fwd\_is \gets initialize(outer\_fwd, inner\_fwd)$ \\
    $bwd\_is \gets initialize(outer\_bwd, inner\_bwd)$ \\
    $state, IG \gets FormatIG(fwd\_is, bwd\_is)$ \\
    \If{state}{
        $AddChildren(root, IG)$ \\
        \If{$IG.num\_nodes > node\_limit$}{
            $DGs \gets Split(fwd\_sg, bwd\_sg)$
            $AddChildren(IG, DGs)$
        }
        $inner\_fwd \gets outer\_fwd$ \\
        $inner\_bwd \gets outer\_bwd$ \\
    }

    $outer\_fwd \gets prev(V^{'}, outer\_fwd)$ \\
    $outer\_bwd \gets next(V^{'}, outer\_bwd)$
}

\Return{root}
\end{algorithm}


    
    
    
    

        

During the process of partitioning, a subgraph tree is constructed to organize the hierarchical structure. As depicted in \autoref{fig:task_tree}, the subgraph tree consists of three distinct levels. The initial level corresponds to the entire DNN graph, representing the largest independent subgraph. The subsequent two levels consist of independent subgraph nodes and dependent subgraph nodes, respectively. Within the tree, divisions are represented by edges, with each non-leaf node having children that denote the corresponding divisions.


When addressing the subgraph tree, the leaf independent subgraphs and dependent subgraphs can be directly optimized. In contrast, the non-leaf nodes require an aggregation process, combining the sub-results from their children based on equations (3) and (9). It is essential to note that the optimization for leaf nodes takes place concurrently to expedite the optimization process.

\subsection{ILP Solver}
We employ Integer Linear Programming (ILP), which is a highly effective method and has been shown to provide near-optimal solutions given enough time, as demonstrated in \cite{ILP1,ILP2,ILP3}. 

The first optimization problem, focused on reducing the theoretical peak memory, can be effectively modeled as an ILP problem. This is achieved by transforming the optimization of operator execution order into a quest for optimizing the lifetimes of tensors, akin to the approach presented in \cite{MODeL}. Each tensor is associated with two pivotal decision variables, denoted as $C$ and $P$, signifying the timesteps of tensor creation and preservation in memory, respectively.

Numerous constraints are incorporated to ensure that the lifetimes of the optimized tensors are related to a valid operator execution order. For instance, tensors generated by the same source must have the same creation timestep and tensors can be created only when all the precedence tensors (inputs of the source operators) are preserved in memory. The theoretical memory is the sum of the corresponding alive tensors size and the objective is minimizing the theoretical peak memory.


For memory layout optimization, tensors are equipped with a decision variable $Offset$, which indicates the address of tensors in the memory space. The most critical constraint in this part is to ensure that tensors with overlapping lifetimes can not have overlapping address spaces, and the target is to minimize the size of the required memory space.

Such ILP methods help deal with complicated memory reuse patterns effectively. Meanwhile, as \autoref{construct_ST} ensures that only fine-grained subgraphs are directly optimized, thereby minimizing the potential impact of the high-complexity ILP method on the efficiency of our approach.

\begin{figure*}[htbp]
    \centering
    \subfloat[Compared with PyTorch]{
        \label{fig:pt_e2e_saving}\includegraphics[width=0.33\textwidth]{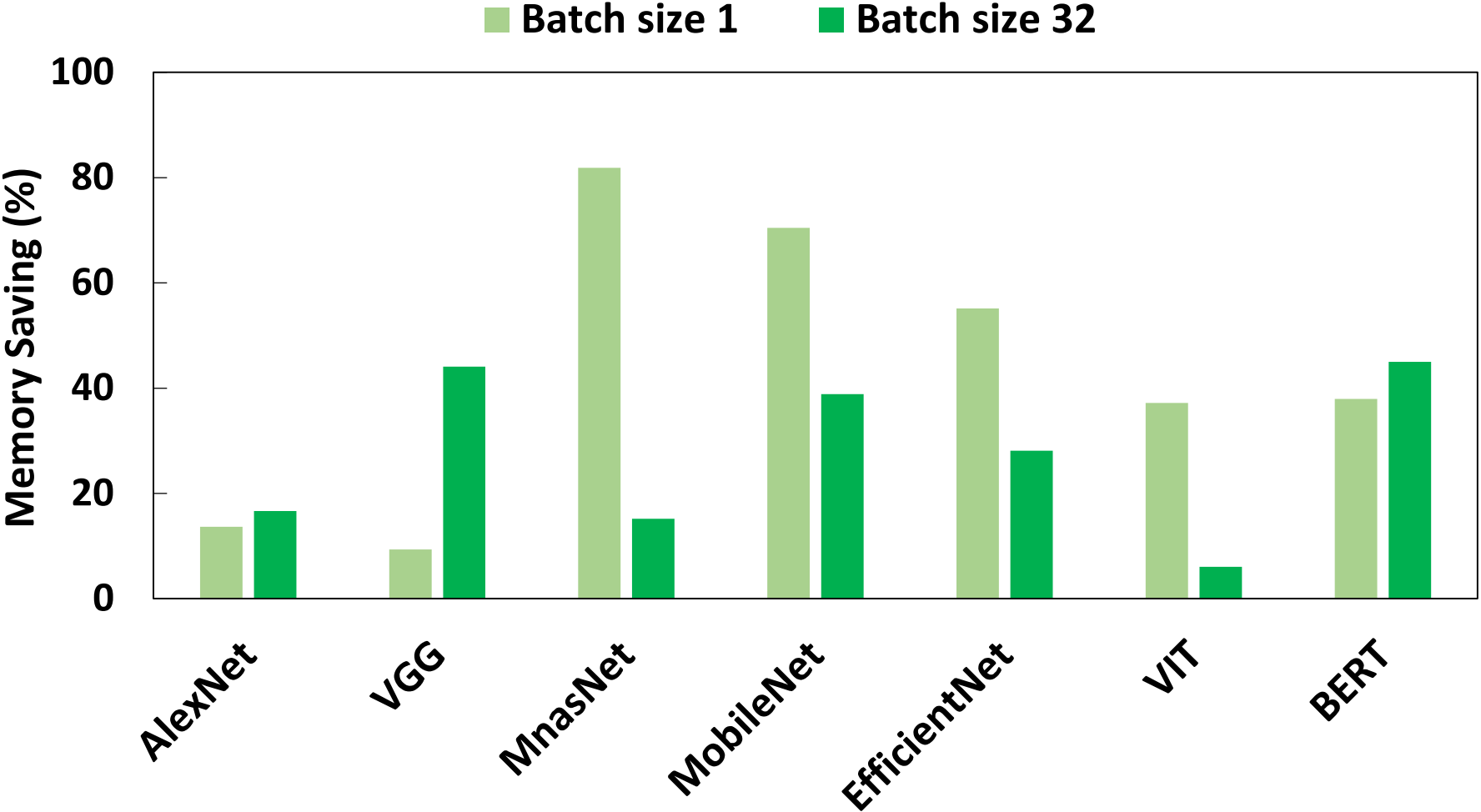}
    }
    \subfloat[Compared with Heuristic (LESCEA+LLFB)]{
        \label{fig:heu_e2e_saving}\includegraphics[width=0.33\textwidth]{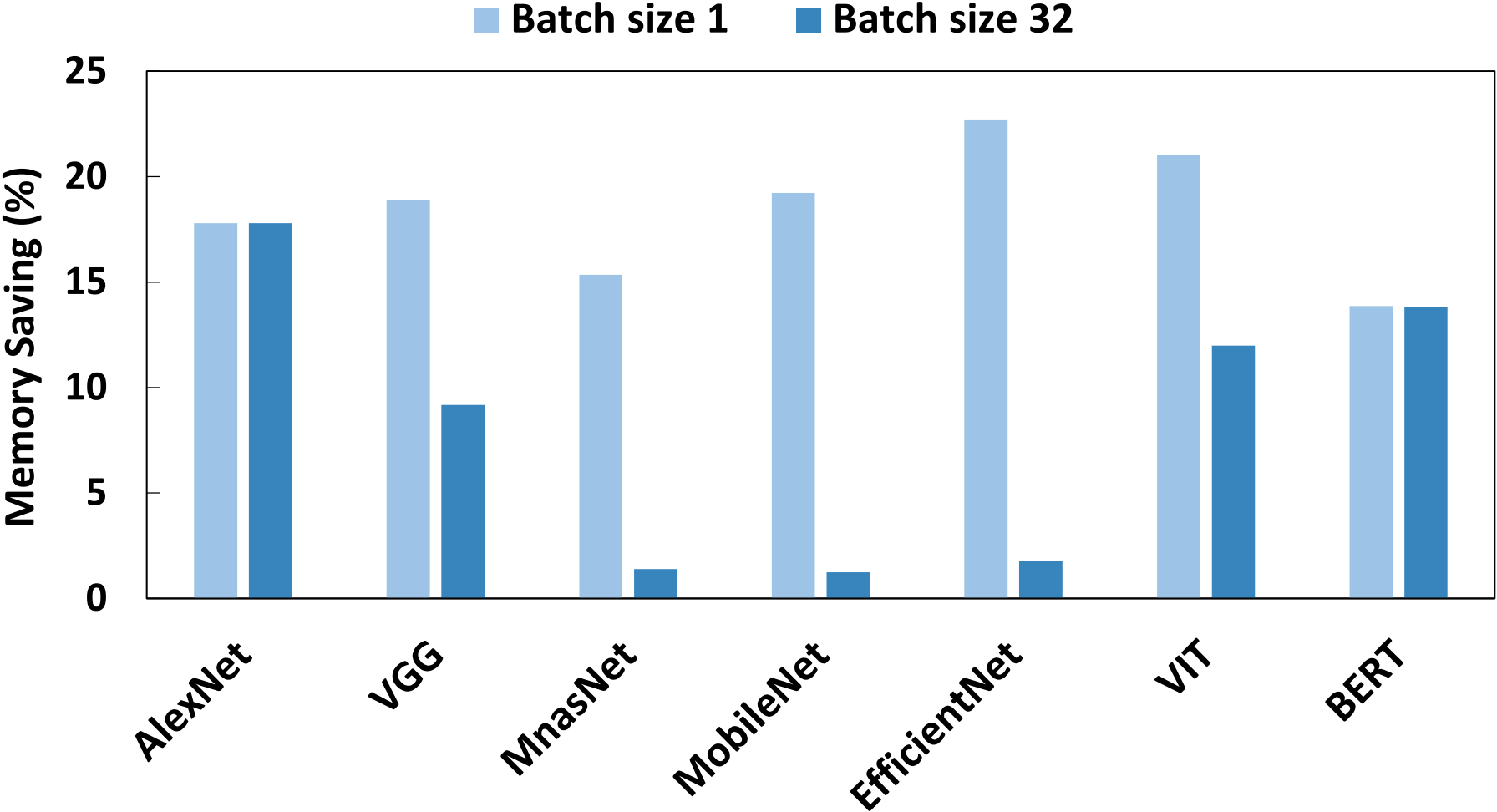}
    }
    \subfloat[Compared with MODeL-Multi-Streaming]{
        \label{fig:MODeL_e2e_saving}\includegraphics[width=0.33\textwidth]{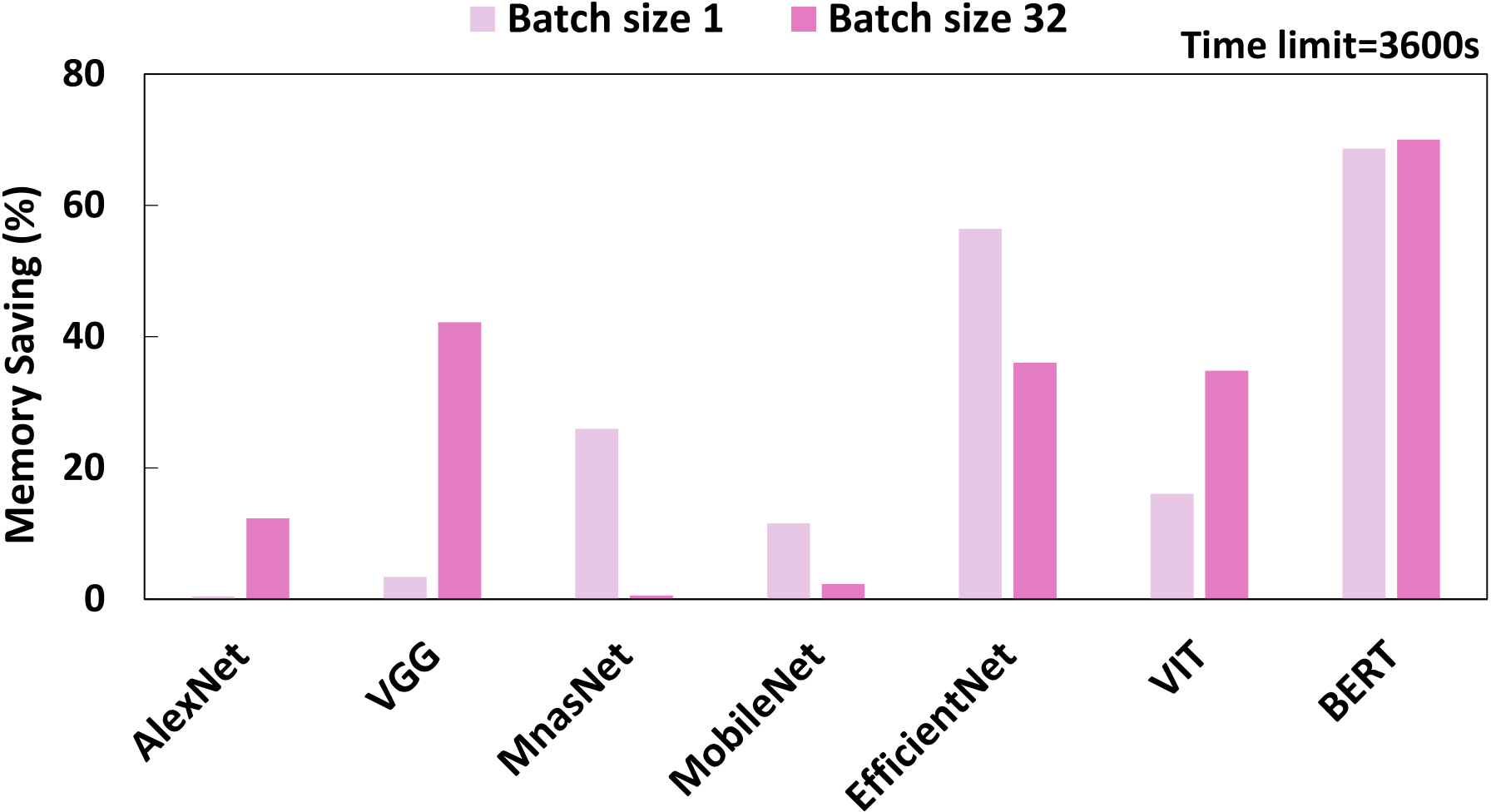}
    }

    \caption{Overall memory reduction (\%) brought by operator execution order and memory layout optimization compared to the baselines.}
    \label{fig:overall_memory_saving}
\vspace{-0.5cm}
\end{figure*}

\section{Evaluation}
In this section, we evaluate \projname from three aspects: memory saving (effectiveness), time-to-optimization (efficiency), and scalability.

\subsection{Experimental setup}
\textit{\textbf{Configurations.}} We implement \projname on the PyTorch v1.13.1 and constructed DNN graphs based on torch.FX. 
All experiments were conducted on a workstation powered by an Intel(R) Xeon(R) Platinum 8269CY CPU operating at 2.50GHz and an NVidia A100 GPU with 80GB of memory.

\par \textit{\textbf{Models.}} We evaluate \projname on the following models:
\begin{itemize}
    \item CNN models: AlexNet\cite{AlexNet}, VGG\cite{VGG}, MnasNet\cite{MnasNet}, MobileNet\cite{MobileNet} and EfficientNet\cite{EfficientNet}. 
    \item Transformer-based models: VIT\cite{VIT}, BERT\cite{BERT}.
    \item Large Language Models: GPT2-XL \cite{GPT2-XL}. 
\end{itemize}

Models except GPT2-XL are evaluated with batch sizes of 1 and 32 because batch size 1 is commonly used when training a model on devices with limited memory capacity, and 32 is used to evaluate memory usage while using a larger input. 
To further demonstrate the practicality of our method, we evaluate the models using \textbf{\textit{Adam}}\cite{Adam} since it is widely used in training and much more complex than other optimizers such as SGD\cite{sgd}.

\par \textit{\textbf{Baselines}}. We compare our method with the following work:
\begin{itemize}
    \item PyTorch. We compare our method with PyTorch, which does not optimize the execution order of operators and the memory layout of tensors. In PyTorch, the operator order is determined by the definition of models in the program, and tensors are assigned dynamically when they are created.
    \item  Heuristic baseline. We select two heuristic methods to format the heuristic baseline. The first one is LESCEA\cite{han2006buffer} which performs scheduling optimization to maximize the buffer sharing in C code generation. Although it is not specifically designed for the DL framework, the prevailing DL compiler XLA\cite{XLA} optimizes the operator execution order with a similar approach which has been proven to have the best performance among all the provided heuristics. Therefore, LESCEA is selected to optimize operator execution order in a heuristic way. The other one is LLFB(\textbf{L}ong-\textbf{L}ived \textbf{F}irst \textbf{B}est-fit) \cite{LLFB}, which has been shown to be equally effective as the ILP method in memory saving for some small instances. 
    \item   ILP-based baseline. Our method is benchmarked against MODeL\cite{MODeL}, a state-of-the-art approach to memory optimization, which employs an Integer Linear Programming (ILP) strategy taking into account both the lifetime and memory offsets of tensors. Notably, MODeL only supports multi-streaming memory optimization, referred to as MODeL-Multi-Streaming (MODeL-MS). To ensure a fair comparison, we implemented our approach under multi-streaming to keep it aligned with the baseline. To make complete evaluations, we also adapted MODeL to facilitate memory optimization for single-streaming operations, resulting in what we call MODeL-Single-Streaming (MODeL-SS). To make this evaluation pragmatic, we enforce a time limit on the MODeL optimization procedure, recognizing that protracted optimization periods can prove counterproductive, essentially negating the very benefit the optimization seeks to provide. 
\end{itemize}

\textit{\textbf{Methodology.}} Generally, The multi-processing implementation and ILP approach may introduce slight fluctuation in time consumption. To minimize the impact of such fluctuation, we conducted 10 benchmark runs and calculated the average. Also, for user-friendliness, we set a time limit of 3600s for ILP solver. 


\subsection{Memory saving evaluation}
We first evaluate the memory consumption on the models mentioned earlier to answer the following questions:
\begin{enumerate}
    \item How effective can our method achieve overall memory saving compared to the baselines?
    \item How effective is our method in operator execution order and memory layout optimization, respectively?
\end{enumerate}

\par To evaluate the effectiveness of our methodology, we compared the overall memory requirements of various neural networks against four baseline approaches. \autoref{fig:overall_memory_saving} illustrates the memory-saving ratio of our approach in comparison to PyTorch, Heuristics (LESCEA+LLFB) and MODeL-Multi-Streaming. The results demonstrate that \projname achieves a remarkable 35.7\%, and 13.3\% reduction in memory usage as compared to PyTorch and heuristics respectively.


\begin{figure*}[t]
    \centering
    \subfloat[Compared with PyTorch]{
        \label{fig:pt_nr_saving}\includegraphics[width=0.33\textwidth]{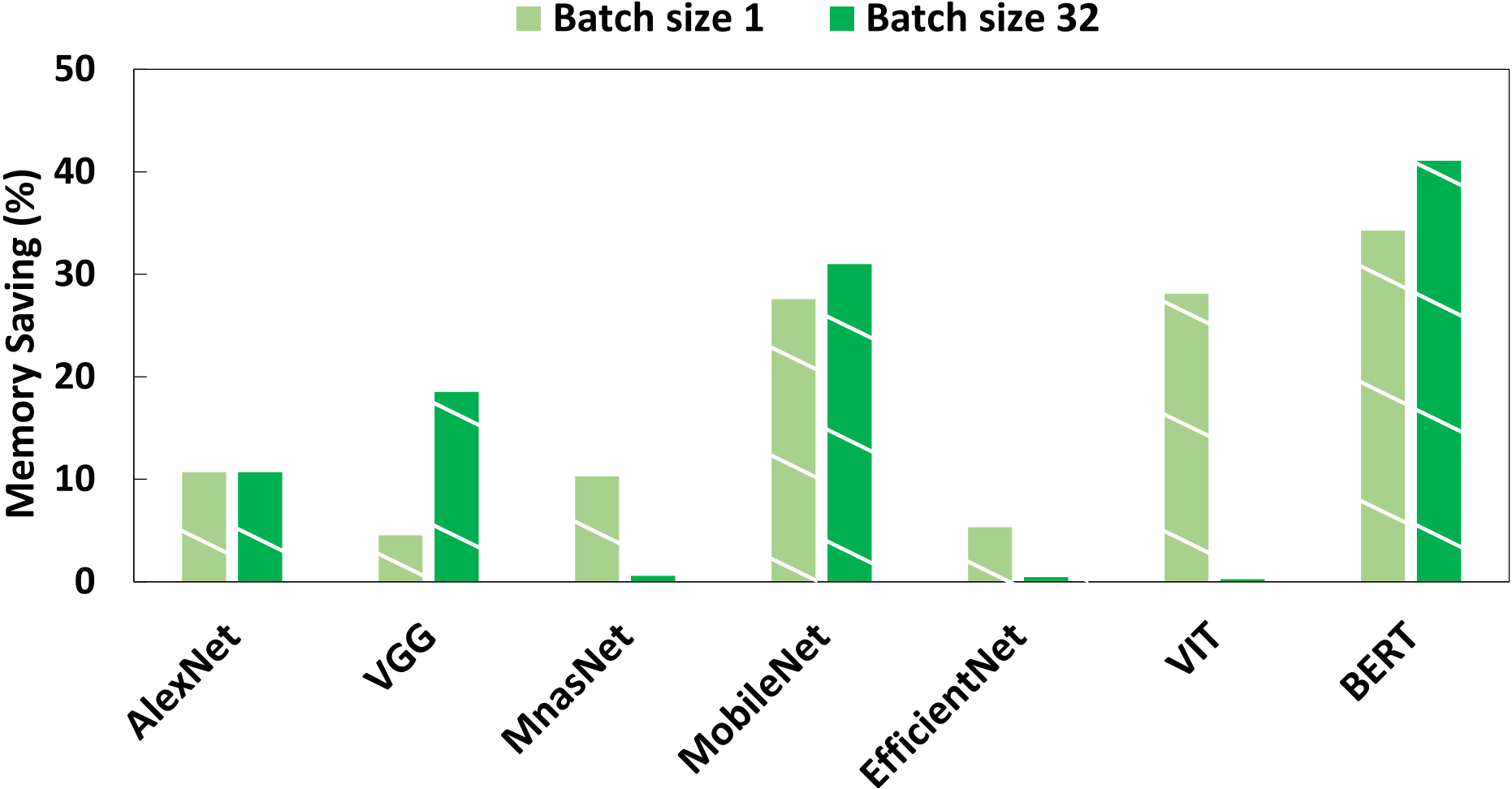}
    }
    \subfloat[Compared with heuristic(LESCEA)]{
        \label{fig:heu_nr_saving}\includegraphics[width=0.33\textwidth]{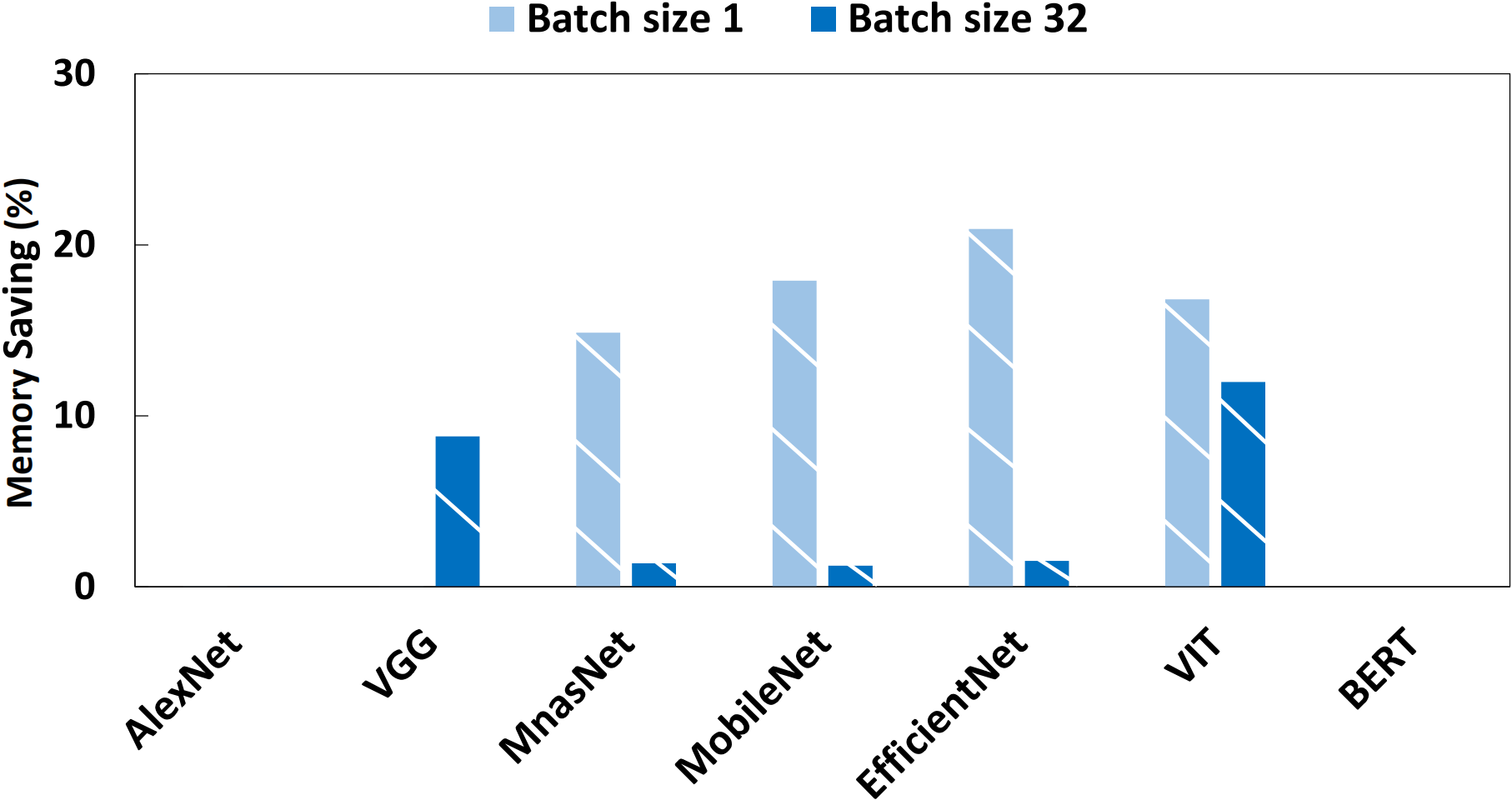}
    } 
    \subfloat[Compared with MODeL-Multi-Streaming]{
        \label{fig:MODeL_ms_nr_saving}\includegraphics[width=0.33\textwidth]{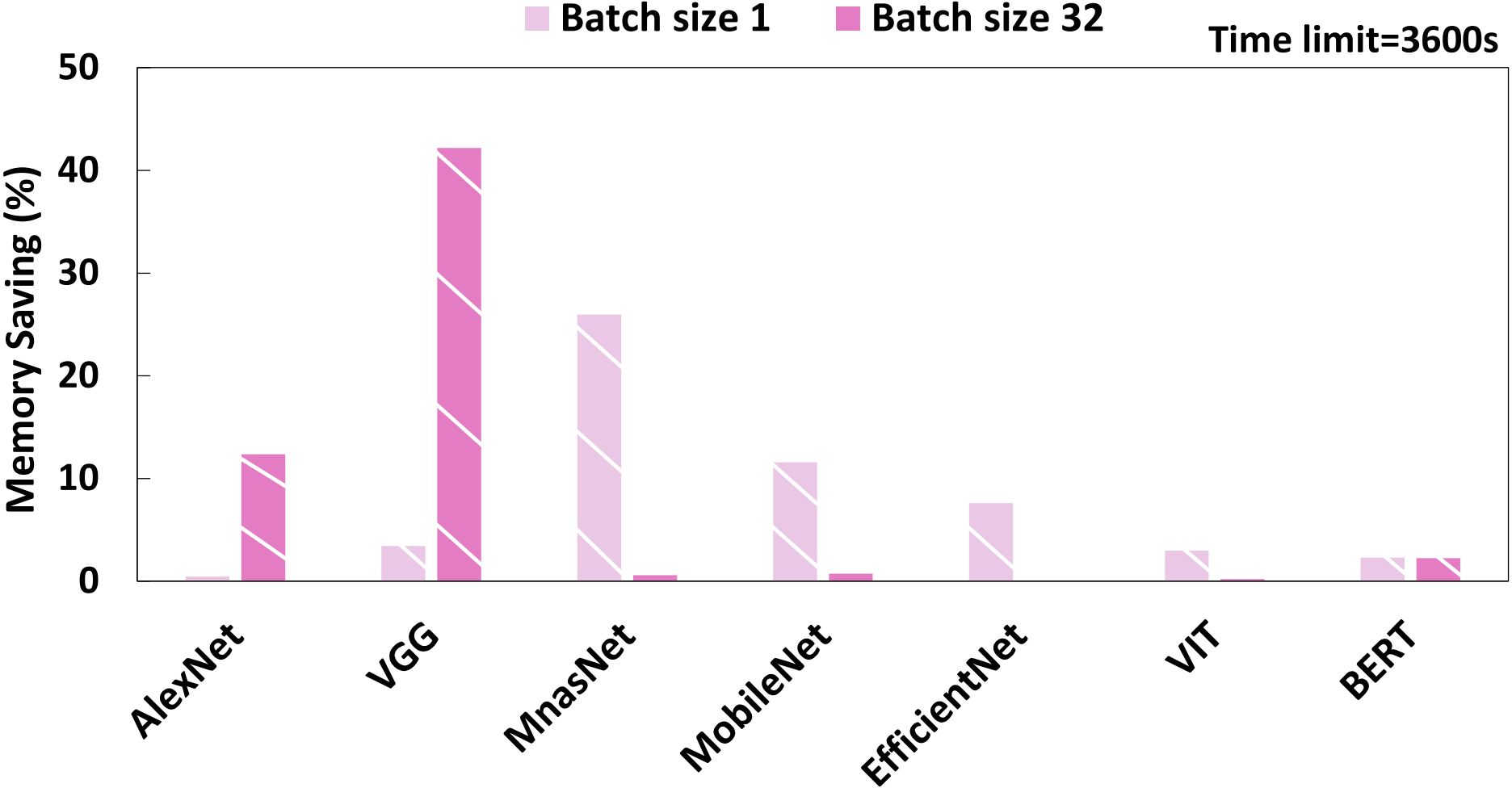}
    }

    \caption{Memory reduction (\%) brought by operator execution order optimization.}
    \label{fig:node_order_mem_saving}
\vspace{-0.6cm}
\end{figure*}

\begin{table}[h]
  \caption{Comparison of fragmentation (\%). For each instance, the top and bottom rows show the results at batch size of 1 and 32.}
  \label{tab:frag}
  \renewcommand{\arraystretch}{1.2} 
  \begin{tabular}{l|cccccc}\hline
      & PyTorch
      & LLFB
      & Ours-SS
      & MODeL-MS
      & Ours-MS \\
     \hline
     \multirow{2}*{AlexNet} & 3.34 & 17.80  & \textbf{0.00} & \textbf{0.00} & \textbf{0.00} \\
     \cline{2-6}
                                                   ~ &  6.65 & 17.80 & \textbf{0.00} & \textbf{0.00} & \textbf{0.00} \\
    \cline{1-6}
    \multirow{2}*{VGG} & 5.05 & 18.89 & \textbf{0.00} & \textbf{0.00} & \textbf{0.00} \\
    \cline{2-6}
                                           ~ & 31.40 & 0.41 & \textbf{0.00 }& \textbf{0.00} & \textbf{0.00} \\
    \cline{1-6}
    \multirow{2}*{MnasNet} & 79.77 & 0.57 & \textbf{0.00} & 0.01 & \textbf{0.00} \\
    \cline{2-6}
                                                    ~ & 14.68 & \textbf{0.00} & \textbf{0.00} & \textbf{0.00} & \textbf{0.00} \\
     \cline{1-6}
     \multirow{2}*{MobileNet} & 59.20 & 1.60 & \textbf{0.00} & \textbf{0.00} & \textbf{0.00} \\
     \cline{2-6}
                                                        ~ & 11.39 & \textbf{0.00} & \textbf{0.00} & 1.59 & \textbf{0.00} \\
     \cline{1-6}
     \multirow{2}*{EfficientNet} & 52.64 & 2.24 & \textbf{0.00} & 52.85 & \textbf{0.04} \\
     \cline{2-6}
                                                           ~ & 27.85 & 0.35 & \textbf{0.00} & 36.04 & \textbf{0.08} \\                  \cline{1-6}             \multirow{2}*{VIT} & 12.63 & 5.08 & \textbf{0.00} & 13.52 & \textbf{0.00} \\
    \cline{2-6}
                                      ~ & 5.82 & \textbf{0.00} & \textbf{0.00} & 34.67 & \textbf{0.00} \\
     \cline{1-6}
     \multirow{2}*{BERT} & 6.97 & 13.86  & \textbf{0.00} & 67.92 & \textbf{0.00} \\
     \cline{2-6}
                                               ~ & 10.54 & 13.86 & \textbf{0.00} & 69.32 & \textbf{0.00} \\
    \hline
\end{tabular}

\vspace{-0.5cm}
\end{table}

For MODeL, we first attempted to modify it to support single-streaming by implementing a constraint that limits the number of operators scheduled at one timestep, which is also implemented in \projname. However, our evaluation revealed that MODeL-Single-Streaming was only capable of providing a solution for Alexnet with a batch size of 1 within the designated time limit of 1h. Optimizing for single-streaming scenarios is more complex than optimizing for multi-streaming scenarios, as the feasible solution space for ILP problems in single-streaming scenarios is a subset of the solution space in multi-streaming scenarios, resulting in more time to find a feasible solution. Therefore, it is hard for MODeL to search for a feasible solution within a user-friendly time in large instances. We allow more time (exceeds 1h) for MODeL to finish the full solving process, and notice that the generated solution is inferior to ours, while our approach can finish the optimization process within 5s, which validates the superiority of our approach.




\par In the case of multi-streaming, as shown in Figure\autoref{fig:MODeL_e2e_saving}, our approach achieves an impressive 27.2\% memory reduction on average when given a time limit 1h. 

\par To further validate the effectiveness of our approach, we evaluated the performance in the specific two aspects: operator execution order and memory layout. On one hand, we record the peak memory required by operators to run these models according to the order. On the other hand, we use fragmentation to measure the effectiveness of memory layout optimization. Fragmentation is defined as the difference between the actual memory requirement and the theoretical peak memory, which is simply the sum of the sizes of the live tensors.

\par \autoref{fig:node_order_mem_saving} shows the impressive relative memory savings that have been achieved via operator order optimization. We find that \projname is capable of yielding up to 41.1\%, 20.9\%, and 42.2\% reduction in theoretical peak memory as compared to PyTorch, LESCEA, and MODeL-Multi-Streaming, respectively. We find that LESCEA struggles to handle scenarios with diverse tensor sizes, leading to suboptimal solutions that are sometimes even worse than PyTorch's native order. While the ILP method can effectively handle scenarios with varying tensor sizes, MODeL faces challenges in addressing weight update scheduling issues, which however can be well supported in \projname. It is worthy of note that the memory reduction for some neural networks in batch size 32 is much smaller than that of batch size 1. We attribute this primarily to the fact that activations' sizes increase with batch size increases, which may reduce the impact of temporary buffers that were previously large. However, there are huge temporary buffers in some cases (e.g., BERT, MobileNet) where our method effectively handles these scenarios.  

\par \autoref{tab:frag} outlines a breakdown of the fragmentation results obtained.  On average, PyTorch exhibits a high fragmentation of 23.0\%, while our proposed approach efficiently controls fragmentation levels to less than 1\% across all tested scenarios. While LLFB can yield 0 fragmentation in some cases, its performance level is unpredictable across all models and may result in fragmentation levels as high as 18.89\%. Our analysis reveals that LLFB struggles to handle cases where tensors' lifetimes are closely intertwined, posing a challenge to achieving effective results in scenarios with a high number of temporary buffers. 

On the other hand, jointly considering the outcomes of  Figure\autoref{fig:MODeL_ms_nr_saving} and Table 1, we deduced that the main reason for MODeL's poor performance in overall memory reduction is its low efficiency in memory layout optimization. It remains unable to complete the memory layout optimization process efficiently for larger models within the context of user-friendly time limits, leading to high fragmentation levels. Although investing more time and resources into optimizing offsets could still produce a final fragmentation outcome of 0, this outcome could require a significantly longer duration, ranging from several days to a significantly extended period which would be obviously non-user-friendly.

\subsection{Time-to-optimization performance evaluation}
In this evaluation, we aim to answer the following two questions:
\begin{enumerate}
\item How does the execution efficiency of our method compare to the baselines? Is the time consumption user-friendly?
\item What is the relationship between the time consumption of our method and the model scale?
\end{enumerate}

\begin{figure}
  \centering
  \includegraphics[width=\linewidth]{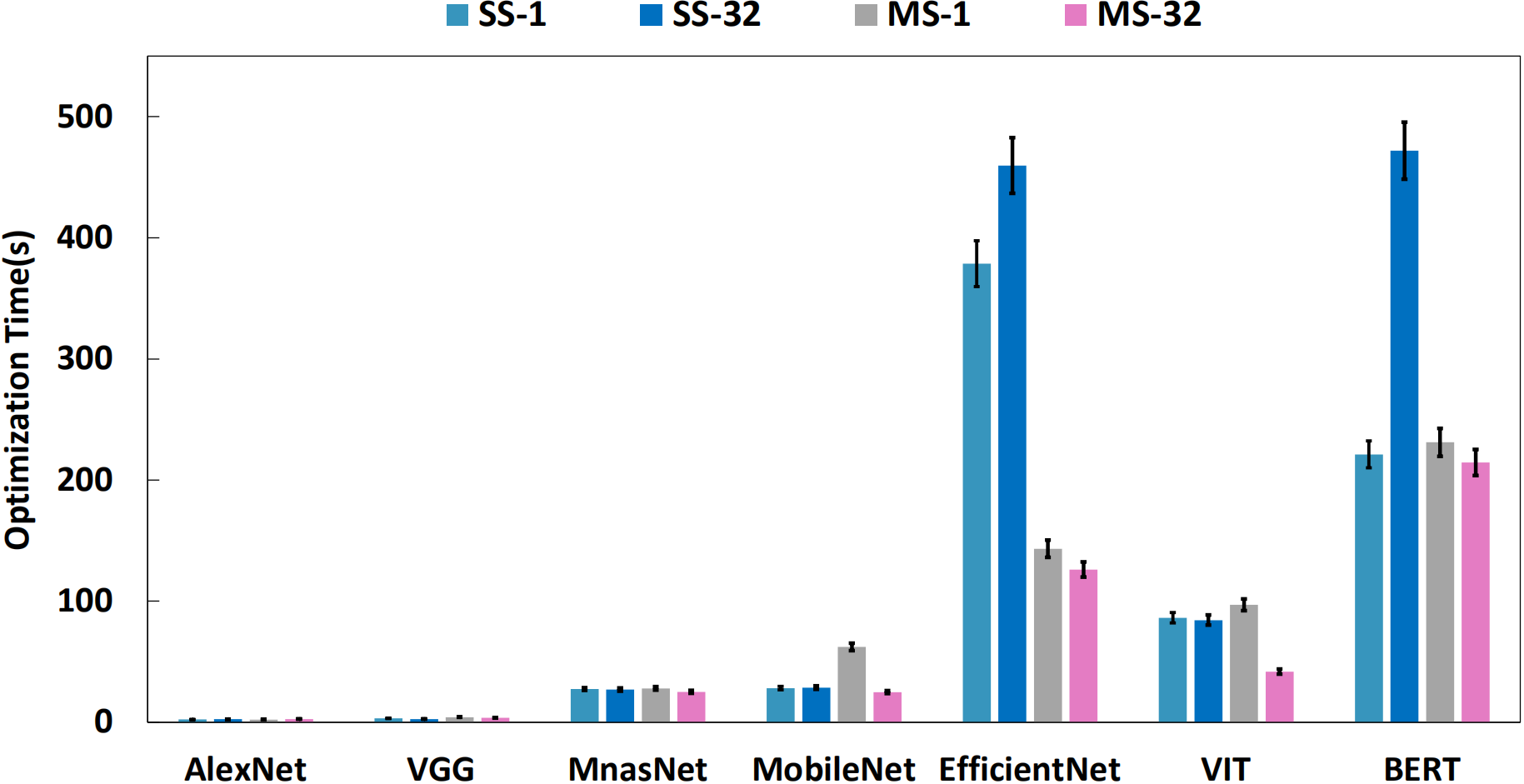}
  \caption{Optimization time of \projname on various models in Single-Streaming (SS) and Multi-Streaming (MS) scenarios.}
  \label{fig:Optimization time}
  \vspace{-0.5cm}
\end{figure}
\begin{figure}
  \centering
  \includegraphics[width=\linewidth]{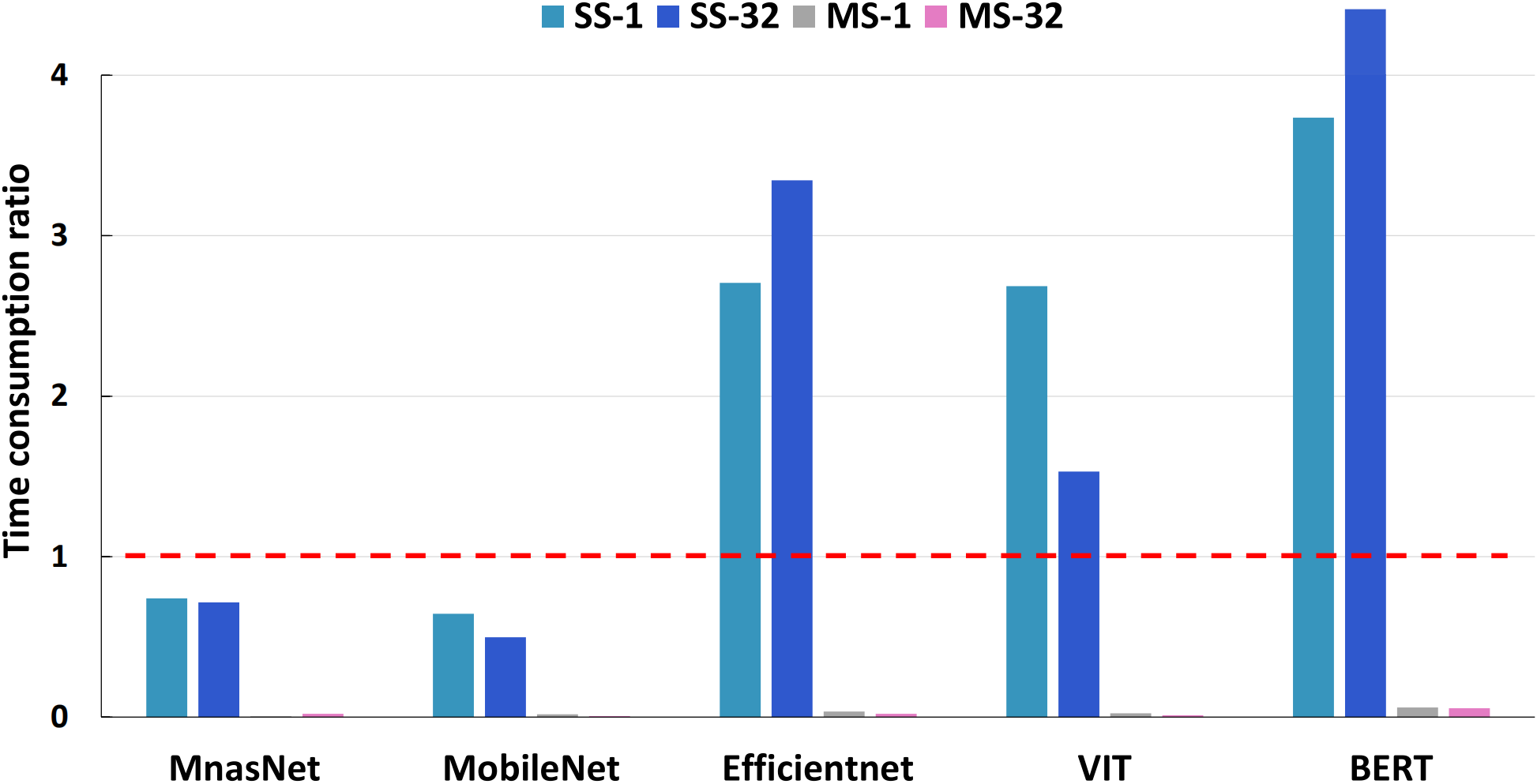}
  \caption{Comparison with Heuristics (LESCEA+LLFB) in SS and MODel in MS for speedup. Since all methods consume very limited time (a few seconds) on AlexNet and VGG, we did not make comparisons between the two models. The ratio is calculated as $\frac{T_{\projname}}{T_{baseline}}$. }
  \label{fig:Time comparison}
\vspace{-0.5cm}
\end{figure}

\par We first measured the time consumed by \projname's. As shown in \autoref{fig:Optimization time}, the optimization of Alexnet and VGG can be completed within 5 seconds, while MnasNet, MobileNet, and VIT can be optimized within or around 100 seconds. While optimization for EfficientNet and BERT takes a relatively longer period, they can still be optimized within 500 seconds. 

\par In order to provide comprehensive evaluation metrics, we have conducted further comparisons with baselines. Specifically, we compared the efficiency of our approach with heuristics (LESCEA+LLFB) in the single-streaming scenario. As MODeL's inability to successfully complete time-effective optimization procedures for most of the models in Single-Streaming, the comparison is rendered non-feasible. Thus, we only conducted a performance comparison against MODeL in the multi-streaming scenario.


As demonstrated in \autoref{fig:Time comparison}, our approach signifies a relatively slower mode of execution compared to that of the heuristics method. Specifically, for the BERT batch size 32 in single streaming, which ate up relatively more time amongst studied models, our procedure requires around 500s whereas heuristics only take roughly 45s. All the same, the optimization time still falls within the tolerable threshold limits, noting that an overall average memory reduction of 13.8\% on average is achievable when analyzed entirely in comparison to heuristics. Furthermore, on both MnasNet and MobileNet models, our method outperformed in obtaining a better solution in a shorter time. On the other hand, \projname outperforms MODeL significantly in terms of speed, achieving at least 53.6x speedup. This demonstrates \projname's ability to always obtain better solutions at a faster rate compared with MODeL.

\begin{figure}
  \centering
  \includegraphics[width=\linewidth]{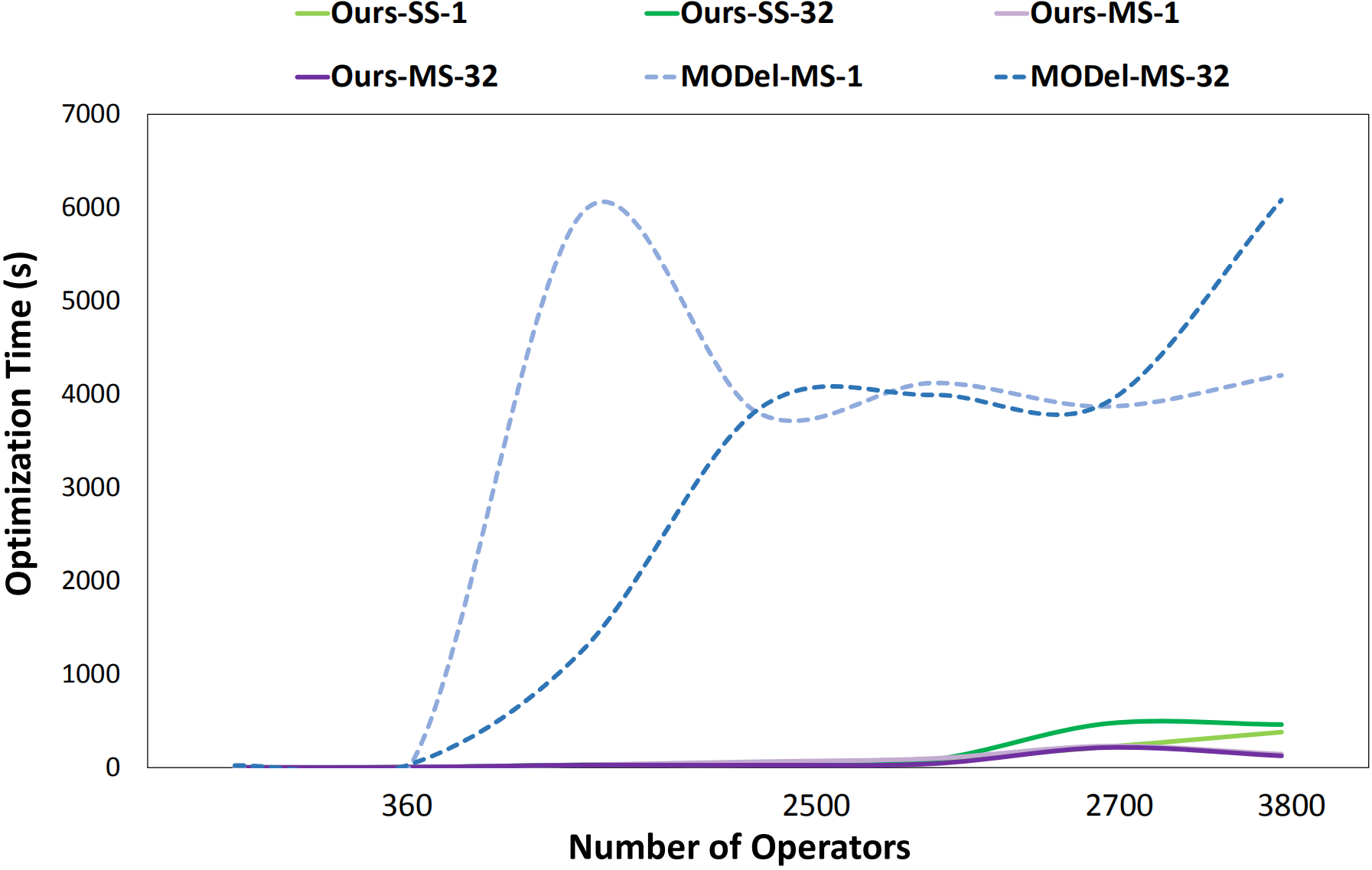}
  \caption{Comparison with MODel in terms of time increasing as operators number increases.}
  \label{fig:op_time}
\vspace{-0.4cm}  
\end{figure}

\begin{figure}[t]
    \centering
    \includegraphics[width=0.45\textwidth]{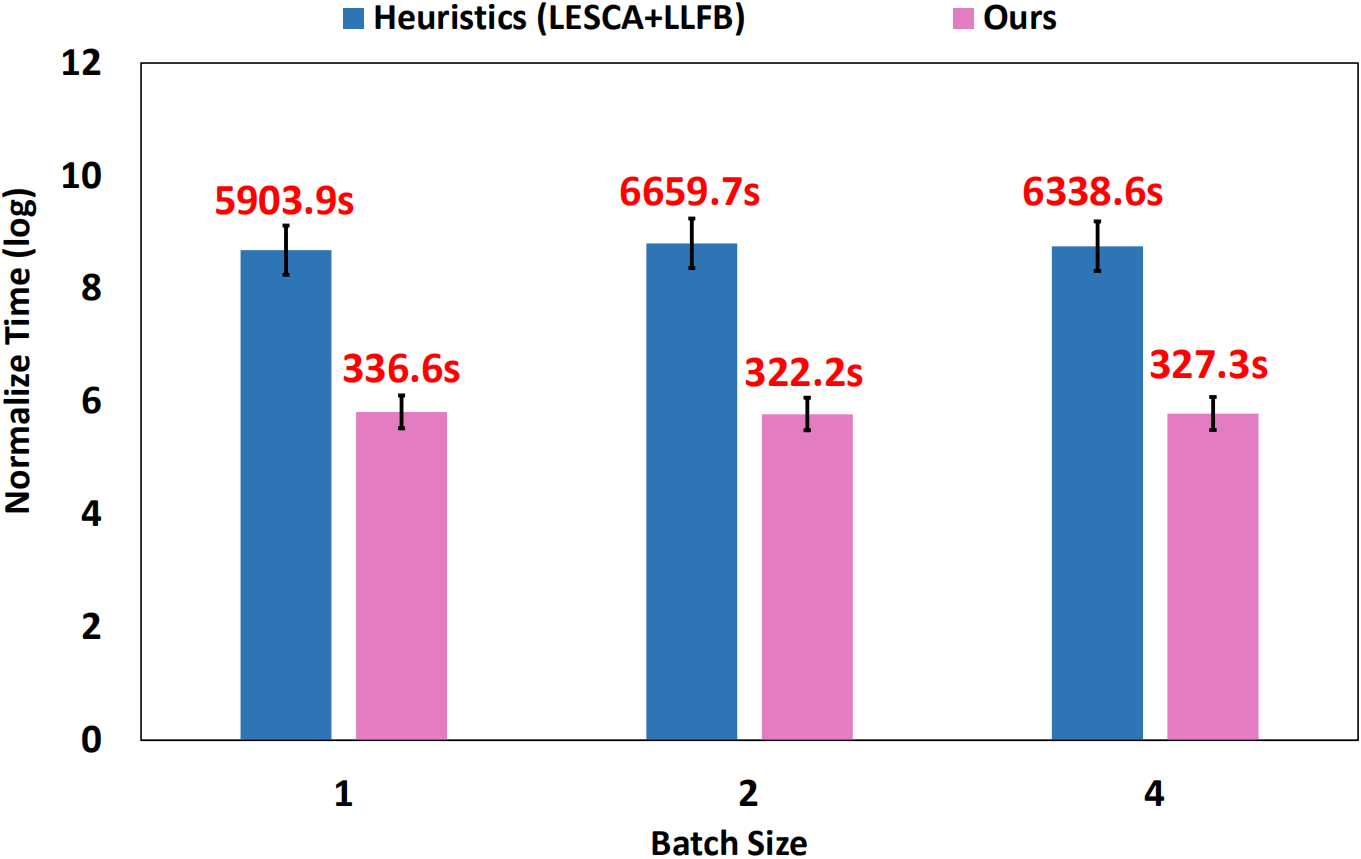}
    \caption{Comparison with Heuristics (LESCEA+LLFB) in terms of the time required to optimize operator execution order and tensor memory layout for GPT2-XL. \projname demonstrates superior speed and efficiency in optimizing large models.
}
    \label{fig:LLM_time}
\vspace{-0.5cm}
\end{figure}
\begin{figure}[ht]
    \centering
    \includegraphics[width=0.45\textwidth]{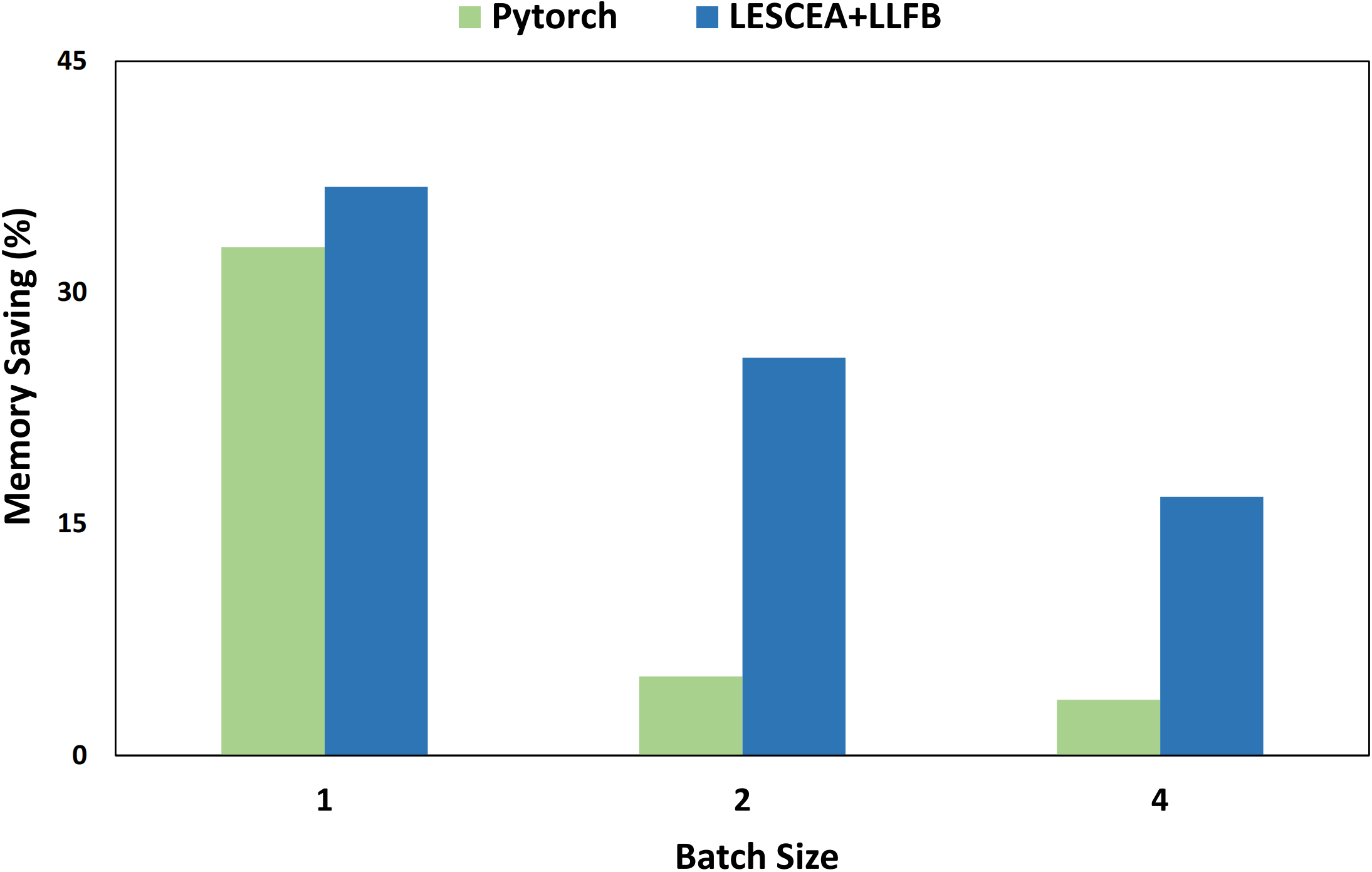}
    \caption{Memory saving of \projname in training GPT2-XL. The top table illustrates the fragmentation of PyTorch, Heuristics (LESCEA+LLFB), and \projname at batch sizes of 1, 2, 4 respectively.}
    \label{fig:LLM_mem}
\vspace{-0.3cm}
\end{figure}

\par The effects of operator number on consuming time are shown in \autoref{fig:op_time}. Despite the fact that the complexity of the ILP solver increases exponentially with an increase in the number of operators, \projname achieves a steady time increase which is primarily caused by the increasing schedule overhead for independent processes. However, we observed a peak at approximately 2700 operators, which corresponds to BERT. This model took longer to optimize compared to other models with more operators due to some large segments in BERT that cannot be further split down. However, the time consumption is still acceptable compared with the state-of-art baseline. 



\subsection{Scalability Evaluation}
Lastly, we conducted further evaluations on the widely used large language model, GPT2-XL, to validate the scalability of our approach. The GPT2-XL graph, with Adam as the optimizer, consists of more than 10 thousand operators, which is substantially larger compared to the earlier-mentioned models.

\par \autoref{fig:LLM_time} illustrates the comparison of time consumption between \projname and heuristics. We see that \projname can optimize GPT2-XL in a similar amount of time as other cases as illustrated in \autoref{fig:Optimization time}, while the optimization time for heuristics significantly increases, leading to reduced usability. It is evident that \projname achieves significantly higher speedup than heuristics for the large case, GPT2-XL. On average, \projname achieves a speedup of 19.2x, which confirms the excellent scalability of \projname.  Furthermore, it is noteworthy that \projname can support the large case that is unsupported in other ILP-based methods. We have attempted to obtain the solutions of MODeL. However, MODeL fails to solve the large ILP model with more than 22 million integer decision variables. 

\par We make comparisons on three batch sizes: 1, 2, 4. \autoref{fig:LLM_mem} shows the overall memory reduction achieved by \projname.  \projname achieves a similar effectiveness compared to heuristics as illustrated earlier. We noticed that for GPT2-XL, PyTorch's execution sequence happened to be better, leading to lower peak memory than heuristic strategies. However, PyTorch's execution sequence depends on the order defined in the program, and it's difficult to ensure the memory efficiency of the execution sequence for all models. It seems that the efficiency of PyTorch and heuristics increases at batch sizes 2 and 4. The reason lies in that the sizes of activations increase incrementally and the temporary buffers occupy a small portion of memory, reducing the optimization space brought by reordering operators and arranging memory layouts. However, for large models, they cannot use larger micro-batch sizes on a single GPU. Therefore, considering the instability of memory-efficiency of PyTorch's execution plan and the availability of small micro-batches on a single GPU, our method remains effective in large model scenarios.


\section{Related Works}

\subsection{High-level mitigation techniques.}
To mitigate memory pressure for GPU, many high-level techniques have been proposed. 
\begin{itemize}
    \item Quantization and Compression.  \cite{low_precision1,low_precision2} propose using reduced precision arithmetic on 16-bit floating point or even quantized representations to train models, significantly reducing memory usage. \cite{COMET,Gist} conducts compression for specific data or layers However, \cite{low_precision_dis} points out that such techniques require careful implementation and can compromise the accuracy of neural networks.
    \item Offloading. \cite{vDNN} virtualize the memory usage of DNNs against both GPU and CPU, moving data out of GPU at the time of memory-intensive. While previous works \cite{SupNeu,caupchin,ZeRO-Offload} are demonstrated to be effective in reducing memory footprint, it increases training time due to extra memory transfers.
    \item Rematerialization. Rematerialization utilizes the extra computation (eg. re-forward propagation) to discard tensors and recompute them when needed. Numerous works \cite{MemOptRec,AutoGC,SupNeu} are proposed to identify tensors to be discarded. However, the introduced computation overhead is still unavoidable. 
\end{itemize}

\subsection{GPU memory management.}
\subsubsection{Operator scheduler Optimization}
The scheduling of operators has been studied to reduce the resource requirements. \cite{schedule} reveals that the scheduling algorithms for the compiled internal representation such as the control/data flow graph (CDFG) have an influence on resource consuming. \cite{OrderInf} also pointed out that the execution order of computations has a potential influence on operator fusion and peak memory usage. However, they have not mentioned how they find an effective execution order. \cite{OrderingChaos} show that the optimal scheduling for directed acyclic graphs is NP-complete. \cite{OrderNPC1, OrderNPC2} optimize the order of operators by traversing the DNN graphs and enumerating all possible topological orders to find a suitable order. However, the complexity remains high even when they leverage various pruning optimizations. As a result, they have only been able to adapt their work for inference on small-scale neural networks. Recently, several works \cite{MODeL} have proposed ILP-based methods for optimizing execution order. However, these methods are limited to simple scenarios. When applied to complex input model structures or high-complexity scenarios, such as single-streaming, their performance is significantly degraded.

Meanwhile, numerous approaches are proposed to execute optimization efficiently. For instance, LESCEA\cite{han2006buffer} introduces a greedy method to schedule operators that result in the least memory increase. However, the theory scheduling the lowest memory increasing operator at every timestep is unable to guarantee the lowest peak memory as it only takes operators' finished state rather than the executing state into consideration. Similarly, XLA\cite{XLA} optimizes the operator execution order heuristically, with a greedy method akin to \cite{han2006buffer} yielding the most favorable results according to their evaluations. 

\subsubsection{Tensors address Optimization} 
Many researchers have explored how to improve memory reuse to reduce fragmentation. Given the NP-complete nature of memory layout optimization as mentioned in \cite{DSA-NP},  it has been extensively studied practically with heuristics. \cite{EfficientMMInf} proposes several greedy approaches based on tensor size and operator breadth (the sum of tensors that are alive during the execution of the operator) to optimize tensor layout in the preallocated buffer. Their methods perform well in inference. However, they only consider the memory reuse pattern in inference, which may not be suitable for training. \cite{LLFB} takes the length of tensors' lifetimes into consideration, placing long-lived tensors on the lowest offset (long-lived-first). They also formulate the memory layout optimization as an ILP problem. Their results show that both the long-lived-first and the ILP method generate similar results for small instances. Since ILP problems can be solved to return nearly optimal results, their heuristic method long-lived-first works very well for small cases. Unfortunately, the solving time of the ILP method becomes unfriendly which makes the optimization problem unsolvable, thus it is not clear how well their system works in large instances. Other ILP-based methods \cite{MODeL, MemoryPlan} show similar disadvantages. They leverage some domain knowledge to simplify the ILP formulation and improve the scalability of the ILP method effectively compared to previous works. However, with the scale of the computation graph increasing, they failed to generate a reasonable solution.

\section{Conclusion}
We present \projname, a DL memory optimization framework that operates on graph level and improves models' memory efficiency by optimizing operator execution order and tensor memory layout. \projname achieves effectiveness, efficiency, and scalability by proposing sophisticated theories and an efficient tree-based algorithm and further enabling the optimization for large models. Experimental results show that \projname significantly reduces the memory consumption and achieves an average speedup of .$53.7\times$ compared to the ILP state-of-the-art method in optimization time. The scalability of \projname has been further validated on GPT2-XL, where it maintains the same effectiveness and efficiency.


\bibliographystyle{ieeetr}
\bibliography{ref.bib}

\end{document}